\documentclass[lettersize,journal]{IEEEtran}
\usepackage{amsmath,amsfonts}
\usepackage{algorithmic}
\usepackage{algorithm}
\usepackage{makecell}
\usepackage{array}
\usepackage{textcomp}
\usepackage{stfloats}
\usepackage{url}
\usepackage{hyperref}
\usepackage{verbatim}
\usepackage{graphicx}
\usepackage{cite}
\usepackage{xcolor}
\usepackage{multicol}
\usepackage{multirow}
\usepackage{tikz}
\usepackage{pgfplots}
\usepackage{pgfplotstable}
\pgfplotsset{compat=1.7}
\usetikzlibrary{patterns}
\usepackage{subcaption}

\begin{document}

\title{RoBERTa-BiLSTM: A Context-Aware Hybrid Model for Sentiment Analysis}

\author{Md. Mostafizer Rahman, Ariful Islam Shiplu, Yutaka Watanobe,~\IEEEmembership{Member,~IEEE}, and Md. Ashad Alam
\thanks{Md. Mostafizer Rahman is with the Department of Computer and Information Systems, University of Aizu, Japan and is with the Information and Communication Technology Cell, Dhaka University of Engineering \& Technology, Gazipur, Bangladesh (e-mail: mostafiz26@gmail.com, mostafiz@duet.ac.bd).}

\thanks{Ariful Islam Shiplu is with the Department of Computer Science and Engineering, Dhaka University of Engineering \& Technology, Gazipur, Bangladesh (e-mail: shipluarifulislam@gmail.com).}

\thanks{Yutaka Watanobe is with the Department of Computer and Information Systems, University of Aizu, Japan (e-mail: yutaka@u-aizu.ac.jp).}

\thanks{ Md. Ashad Alam is with the  Ochsner Center for Outcomes Research, New Orleans, LA 70121, USA (e-mail: mdashad.alam@ochsner.org).}

\thanks{\copyright 2025 IEEE.  Personal use of this material is permitted.  Permission from IEEE must be obtained for all other uses, in any current or future media, including reprinting/republishing this material for advertising or promotional purposes, creating new collective works, for resale or redistribution to servers or lists, or reuse of any copyrighted component of this work in other works.}

}

\markboth{Journal of \LaTeX\ Class Files,~Vol.~14, No.~8, August~2021}%
{Shell \MakeLowercase{\textit{et al.}}: A Sample Article Using IEEEtran.cls for IEEE Journals}


\maketitle

\begin{abstract}
With the rapid advancement of technology and its easy accessibility, online activity has become an integral part of everyday human life. Expressing opinions, providing feedback, and sharing feelings by commenting on various platforms, including social media, education, business, entertainment, and sports, has become a common phenomenon. Effectively analyzing these comments to uncover latent intentions holds immense value in making strategic decisions across various domains. However, several challenges hinder the process of sentiment analysis including the lexical diversity exhibited in comments, the presence of long dependencies within the text, encountering unknown symbols and words, and dealing with imbalanced datasets. Moreover, existing sentiment analysis tasks mostly leveraged sequential models to encode the long dependent texts and it requires longer execution time as it processes the text sequentially. In contrast, the Transformer requires less execution time due to its parallel processing nature. In this work, we introduce a novel hybrid deep learning model, RoBERTa-BiLSTM, which combines the Robustly Optimized BERT Pretraining Approach (RoBERTa) with Bidirectional Long Short-Term Memory (BiLSTM) networks. RoBERTa is utilized to generate meaningful word embedding vectors, while BiLSTM effectively captures the contextual semantics of long-dependent texts. The RoBERTa-BiLSTM hybrid model leverages the strengths of both sequential and Transformer models to enhance performance in sentiment analysis. We conducted experiments using datasets from IMDb, Twitter US Airline, and Sentiment140 to evaluate the proposed model against existing state-of-the-art methods. Our experimental findings demonstrate that the RoBERTa-BiLSTM model surpasses baseline models (e.g., BERT, RoBERTa-base, RoBERTa-GRU, and RoBERTa-LSTM), achieving accuracies of 80.74\%, 92.36\%, and 82.25\% on the Twitter US Airline, IMDb, and Sentiment140 datasets, respectively. Additionally, the model achieves F1-scores of 80.73\%, 92.35\%, and 82.25\% on the same datasets, respectively. 
\end{abstract}


\section{Introduction} \label{sec:introduction}
\IEEEPARstart{I}{n} today's world, technological advancements have empowered individuals to provide feedback and review, express opinions, and share feelings across various platforms such as social media, entertainment, education, programming, sports, and business. Particularly, social media platforms have emerged as primary mediums for communication, facilitating discussions on a broad spectrum of topics \cite{jothi2011analysis, bruns2015habermas, xiong2014opinion, oliveira2020people}. The comments generated on these platforms hold significant value for decision-making and strategy formulation. Sentiment analysis\footnote{Sentiment Analysis, Comment Analysis, and Text Analysis are used interchangeably to convey the same meaning.}, the process of extracting actual sentiment or underlying meaning from comments \cite{pandey2023bert}, plays a crucial role in understanding public thinking. It has emerged as a prominent topic in the field of Natural Language Processing (NLP) due to its significance \cite{tam2021convbilstm}. Understanding the latent intentions within user comments is crucial for various applications, including brand monitoring \cite{tedeschi2015cloud}, market analysis \cite{rao2012analyzing, bharathi2017sentiment}, and sentiment analysis \cite{alrumaih2020sentiment}. However, accurately discerning the intended meaning behind comments remains a considerable challenge. Hence, it is crucial to devise an effective sentiment analysis model capable of comprehending long-distance dependencies, unfamiliar words and symbols, as well as code-mixed languages (comments containing two or more languages), while also adeptly managing the lexical diversity present in texts.


A significant number of approaches have been proposed employing machine learning (ML), deep learning (DL), Recurrent Neural Networks (RNNs), and Transformer \cite{vaswani2017attention}-based large language models (LLMs) for comment analysis. In a study \cite{wongkar2019sentiment}, three ML models—Naïve Bayes (NB), Support Vector Machine (SVM), and K-Nearest Neighbor (KNN)—were applied to Twitter data to comprehend people's sentiments. The NB model achieved an accuracy of 75.58\% and outperformed the other models. In \cite{saad2020opinion}, Logistic Regression (LR), XgBoost, NB, Decision Tree (DT), SVM, and Random Forest (RF) ML algorithms were utilized for sentiment analysis of Twitter US Airline dataset. Text preprocessing steps, such as stop word and punctuation removal, case folding, and stemming, were performed before model training. The SVM model achieved an accuracy of 83.31\%, surpassing the performance of the compared models. Similarly, various ML algorithms have been employed to conduct sentiment analysis on diverse datasets such as Sentiment140 and Airline reviews \cite{prabhakar2019sentiment, rahman2019efficient,madhuri2019machine,rahat2019comparison,jung2016enhanced}. 

In addition to ML algorithms, RNN models like Long Short-Term Memory (LSTM), Bidirectional LSTM (BiLSTM), Gated Recurrent Unit (GRU), and Convolutional Neural Network (CNN) have been utilized, achieving superior accuracy for comment analysis compared to ML models \cite{uddin2019depression, rahman2020source,alahmary2019sentiment,dholpuria2018sentiment,thinh2019sentiment}. Surveys of DL models for sentiment analysis presented in research \cite{dl_survey_9260162,dl_survey_9085334}. In some studies, researchers have explored techniques utilizing pre-trained word embeddings such as Doc2vec, Word2vec, fastText, and GloVe \cite{tan2023roberta}, while word embedding is crucial for sentiment analysis. Many ML and DL models have been proposed for sentiment analysis, with sequential models particularly adept at encoding long-distance dependencies in text. However, sequential models are computationally less efficient due to their serialized processing capability. In contrast, Transformer-based LLMs take comparatively less computational time due to their parallelized processing capability.

In recent years, advancements in LLMs, particularly Transformer-based architectures such as Generative Pre-trained Transformer (GPT) \cite{radford2018improving}, Bidirectional Encoder Representations from Transformers (BERT) \cite{devlin2018bert}, and Robustly Optimized BERT Pretraining Approach (RoBERTa) \cite{liu2019roberta}, have offered even greater potential for improving sentiment analysis tasks \cite{wu2021bert,gao2019target,alaparthi2021bert,wu2024research,cheruku2023sentiment,liao2021improved}. The Transformer-based model leverages the attention mechanism \cite{bahdanau2014neural} which makes it more effective in NLP tasks. Particularly, attention mechanism calculates a weighted sum of the input embeddings, where the weights are determined by a learned compatibility function between the \textit{query} and \textit{key} embeddings \cite{ tan2023roberta}. This capability empowers the model to adeptly capture long-range dependencies within the input sequence, resulting in the creation of more informative representations. Younas et al. \cite{younas2020sentiment} proposed two LLMs, Multilingual BERT (mBERT) and XML-RoBERTa (XML-R), for the analysis of code-mixed language comments on a Twitter dataset. Experimental results demonstrated that mBERT and XML-R achieved accuracy ($\mathbf{A}$) scores of 69\% and 71\%, respectively. In a study \cite{dhola2021comparative}, the BERT model exhibited superior performance (with an $\mathbf{A}$ of 85.4\%) compared to ML models for comment analysis. Moreover, comprehensive surveys on text analysis with Transformer-based LLMs are presented in studies \cite{sur_9996141,sur_10380590}. Poria et al. \cite{poria2020beneath} discussed existing challenges and explored new research directions in sentiment analysis.


In this paper, we propose a hybrid model, RoBERTa-BiLSTM, designed for sentiment analysis. This model harnesses the strengths of both Transformer-based LLM, RoBERTa, and RNN-based model, BiLSTM. The RoBERTa model serves as the encoder, tokenizing the input sequence and encoding it into representative word embeddings. These embeddings are then fed into the BiLSTM via a dropout layer. The BiLSTM model effectively captures the long-range dependencies within the word embeddings while mitigating the gradient vanishing issue often encountered in RNNs. By processing the input sequence in both forward and backward directions, the BiLSTM enhances the model's contextual understanding of the text \cite{rahman2023multilingual, rahman2021bidirectional}. To discern the relationship between the output of the BiLSTM model and the class labels, a dense layer is incorporated. Additionally, a Softmax function is applied to the classification layer to estimate the probability distribution of the class labels. The main contributions of the paper are outlined as follows:

\begin{enumerate}

    \item [(i)] We propose a hybrid context-aware model, RoBERTa-BiLSTM, for sentiment analysis. This model combines the strengths of LLM and RNN to enhance the understanding of textual content. The RoBERTa model is employed to generate representative word embeddings of the text. RoBERTa's features, including extensive training with large datasets, training on longer sequences, removal of the next sentence predictive objective, and dynamically changing masking patterns during training, render it highly effective in sentiment analysis tasks. Conversely, the bidirectional data processing nature of BiLSTM aids in capturing long-range temporal dependencies within word embeddings, which is beneficial in text analytics.

    \item[(ii)] Experimental results reveal that RoBERTa-BiLSTM achieved an $\mathbf{A}$ of 80.74\% for Twitter US Airline, 92.36\% for IMDb, and 82.25\% for  Sentiment140 datasets. These findings demonstrate that RoBERTa-BiLSTM outperforms RoBERTa-base, RobERTa-GRU, RobERTa-LSTM, and other state-of-the-art models.

    \item[(iii)] Hyperparameters are fine-tuned to ascertain the optimal parameters for the RoBERTa-BiLSTM model for sentiment analysis. Data augmentation is applied to an imbalanced dataset to showcase the model's performance before and after augmentation.

\end{enumerate}

The rest of the paper is organized as follows: Section \ref{relatedworks} presents a review of related research employing ML, DL, and LLMs for sentiment analysis. Section \ref{task_description} presents the task description of this research. Section \ref{proposedapproach} elaborates the proposed RoBERTa-BiLSTM approach for sentiment analysis. Section \ref{dataset} presents details on the dataset and the preprocessing steps undertaken for experiments. Section \ref{hyper_tuning} provides insight into the hyperparameters employed and their fine-tuning policy in this research. Following that, in Section \ref{experimental_results}, we present comprehensive experimental results spanning various datasets to showcase the performance of the model. Section \ref{discussion} engages in a discussion of the obtained results. Lastly, in Section \ref{conclusion_research}, we conclude this study, reflecting on its findings and offering insights into future research directions.

\section{Related Works} \label{relatedworks}


In this section, we present a comprehensive review of sentiment analysis research, focusing on various methods employed, including ML, DL, and LLM. Additionally, our literature review encompasses diverse datasets, such as movie reviews, social media text, program code, mixed text, airline review text, and Covid-19 datasets. Through an in-depth analysis, we explore the effectiveness and applicability of different techniques and models for sentiment analysis across diverse domains.

\subsection{Machine Learning}

The study \cite{wongkar2019sentiment} contributes to understanding public sentiment towards 2019 Republic of Indonesia presidential candidates by conducting sentiment analysis on Twitter data. Three ML algorithms (e.g., NB, SVM, and KNN) are used to classify sentiments, providing insights into public opinion dynamics during the election period. Several steps including data collection and text preprocessing are taken into account before the training process. The results indicate that the sentiment polarity of the combined data achieves an $\mathbf{A}$ of 80.1\%.
Rahat et al. \cite{rahat2019comparison}  discuss the importance of sentiment analysis across various domains and propose the development of a platform to discern opinions as positive, negative, or neutral using supervised ML techniques. It highlights data preprocessing techniques applied to social media content for extracting structured reviews. Additionally, it explores the application of algorithms to classify sentiments. The results of the experiment indicate that SVM outperforms NB algorithm in terms of overall $\mathbf{A}$, precision ($\mathbf{P}$), and recall ($\mathbf{R}$) values, particularly in the context of analyzing airline reviews. The evaluation demonstrates the effectiveness of the proposed approach in accurately classifying sentiments across different domains. 

The study \cite{jung2016enhanced} utilizes Sentiment140 for computer simulations, a widely used dataset for sentiment analysis tasks due to its large volume of Twitter messages annotated with sentiment labels. The contribution of this paper lies in its proposition of a novel scheme for sentiment analysis, tailored for real-time stream data, by integrating Laplace Smoothing with Binarized Naive Bayes Classifier (NBC) and leveraging the distributed and parallel processing capabilities of SparkR. In terms of accuracy, the computer simulations conducted with Sentiment140 consistently demonstrate superior performance of the proposed approach over existing schemes. The integration of NBC effectively improves sentiment analysis accuracy, while the utilization of SparkR environment further enhances efficiency, making real-time sentiment analysis feasible and accurate. Madhuri and collaborators \cite{madhuri2019machine} focus on collecting tweets. This targeted tweets dataset enables the framework to provide contextually relevant sentiment analysis for the railway domain. They proposed a framework that employs a comprehensive evaluation procedure. This empirical study also demonstrates the effectiveness of the proposed framework in sentiment analysis. Through rigorous evaluation in terms of $\mathbf{P}$, $\mathbf{R}$, and F1-score ($\mathbf{F1}$), the framework showcases its utility in extracting valuable insights from social media data, thereby facilitating informed decision-making and strategic planning for enterprises operating in the railway domain. Prabhakar et al. \cite{prabhakar2019sentiment} use different places on the internet to collect what customers think about US Airlines. This includes places like Skytrax where people leave reviews and Twitter where they share short messages. These places give important information used to teach and test the new way of understanding people's feelings. Their research introduces an improved Adaboost approach for sentiment analysis, which obtained the highest $\mathbf{P}$, $\mathbf{R}$, and $\mathbf{F1}$ scores of 78\%, 65\%, 68\%, respectively. 

Saad \cite{saad2020opinion} leveraging a dataset comprising tweets from different US airlines. The process begins with preprocessing steps, including cleaning the tweets and extracting features to create a Bag of Words (BoW) model. This paper employs six ML algorithms—SVM, LR, RF, XgBoost, NB, and DT—in the classification phase to categorize tweets. The author utilizes the K-Fold Cross-Validation technique to split the data into 70\% of training and 30\% of testing for validation purposes. The accuracy of each classifier is evaluated using metrics such as $\mathbf{A}$, $\mathbf{P}$, $\mathbf{R}$, and $\mathbf{F1}$. After comparing the results, SVM emerges with the highest $\mathbf{A}$ of 83.31\%, signifying its effectiveness in categorizing tweets into sentiment categories. 
Shiplu et al. \cite{shiplu2023robust} effectively address the critical need for precise comment classification in the era of rapid data expansion. Their comprehensive methodology integrates diverse machine learning algorithms and voting techniques, yielding promising results. The proposed ensemble model, \textit{RF+AdaBoost+SVM+Soft-Voting}, attains an impressive $\mathbf{A}$ of approximately 98\% on a YouTube dataset.

\subsection{Deep Learning}

In recent years, researchers have made significant advancements in sentiment analysis by employing DL techniques. Rhanoui et al. \cite{rhanoui2019cnn} proposed a CNN-BiLSTM model with Doc2vec embeddings, achieving a remarkable $\mathbf{A}$ of 90.66\%, surpassing existing methods. Similarly, Anbukkarasi et al. tailored a combined character-based DBLSTM model for sentiment analysis of Tamil tweets, outperforming LSTM with an $\mathbf{A}$ of 86.2\% and an $\mathbf{F1}$ of 81\% \cite{anbukkarasi2020analyzing}. Dholpuria et al.  \cite{dholpuria2018sentiment} conducted a comparative analysis between DL and traditional supervised ML classifiers for sentiment analysis of movie reviews. They found that combining DL with traditional methods significantly improved classification accuracy, with CNN achieving an $\mathbf{A}$, $\mathbf{P}$, $\mathbf{R}$, and $\mathbf{F1}$ of 99.33\%, 99.67\%, 99.02\%, and 99.35\%, respectively. Alahmary et al. \cite{alahmary2019sentiment} filled a research gap by applying DL techniques to sentiment analysis of Saudi dialect texts, with BiLSTM achieving the highest  $\mathbf{A}$ of 94\%, surpassing LSTM ( $\mathbf{A}$ of 92\%) and SVM ( $\mathbf{A}$ of 86.4\%). Thinh et al. \cite{thinh2019sentiment} utilized DL models with a feature extractor comprising convolutional, max pooling, and batch normalization layers, achieving promising results with an $\mathbf{A}$ of 90.02\% on the IMDb review sentiment dataset. Haoyue et al. \cite{dlsurvey9260162} conducted a survey on DL methods for sentiment analysis. They comprehensively discuss benchmark datasets, evaluation metrics, and the performance of existing DL methods. Furthermore, they address current challenges and future research directions in sentiment analysis using DL methods. These studies collectively demonstrate the effectiveness of DL approaches in accurately discerning sentiments from textual data across diverse domains and languages.

\begin{figure*} [h]
\centering
  \includegraphics[width=.85\linewidth]{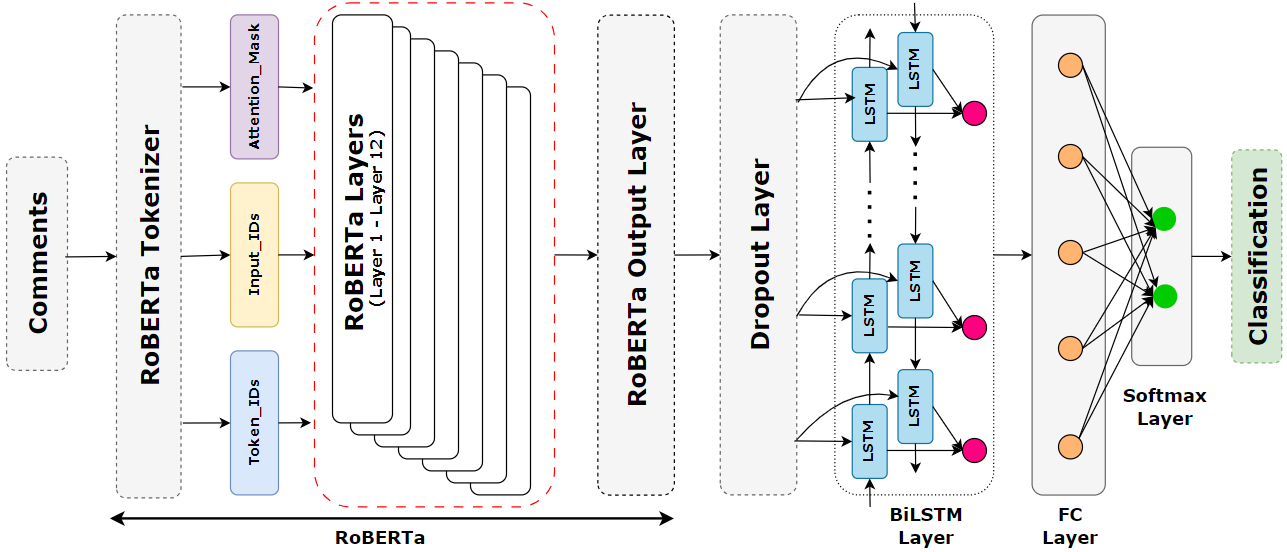}
  \caption{The proposed RoBERTa-BiLSTM hybrid model architecture}
  \label{proposed_architecture}
\end{figure*}

\subsection{Large Language Model}

Singh et al. \cite{singh2021sentiment} employed the BERT model for sentiment analysis of COVID-19 datasets, achieving a notable $\mathbf{A}$ of 94\% on tweets collected from various regions. Tan et al. \cite{tan2022roberta} presents a comprehensive approach to sentiment analysis in social media text, addressing the challenges of lexical diversity and imbalanced datasets. They introduce innovative data augmentation techniques using GloVe word embeddings and propose a hybrid DL model that integrates RoBERTa and LSTM. Their experimental results demonstrate the effectiveness of the hybrid model, which surpasses state-of-the-art methods across multiple datasets. Their study highlights the potential practical applications of their approach in sentiment analysis, particularly in the context of social media data. The study \cite{younas2020sentiment} contributes to the field by addressing the challenge of analyzing imperfect and informal languages, such as code-mixed Roman Urdu and English, prevalent on social media platforms. They propose a state-of-the-art DL model for sentiment analysis, leveraging mBERT and XLM-R models. Experimental results demonstrate that the XLM-R model, with tuned hyperparameters, outperforms mBERT, achieving an $\mathbf{F1}$ score of 71\%. This research underscores the efficacy of the XLM-R model in handling code-mixed text sentiment analysis without relying on lexical normalization or language dictionaries, thereby contributing significantly to the advancement of sentiment analysis techniques for diverse linguistic contexts. Furthermore, pretrained LLMs are being utilized for sentiment analysis across various domains, including education, film, energy, feature unification, feature extraction, and more \cite{10478509,9200457,9466122,10231319,10458136,10284535}.


\section{Task Description} \label{task_description}

Let $x=\{x_1, x_2, x_3, \cdots x_n\}$ be the set of input text. The input text can be represented as an embedding matrix, with its mathematical formulation described by Equation (\ref{embedding_matrix}).

\begin{equation} \label{embedding_matrix}
E_{l,d} = 
\begin{Bmatrix}
e_{1,1} & e_{1,2} & \cdots & e_{1,d} \\
e_{2,1} & e_{2,2} & \cdots & e_{2,d} \\
\vdots  & \vdots  & \ddots & \vdots  \\
e_{l-1,1} & e_{l-1,2} & \cdots & e_{l-1,d} \\
e_{l,1} & e_{l,2} & \cdots & e_{l,d} 
\end{Bmatrix}
\end{equation}

where $E  \in \mathbb{R} ^{l \times d}$ is the embedding matrix, $d$ is the dimension of word embedding, and $l$ is the text length. Each word ($\alpha$) in the text ($x$) can be represented as a $d$-dimensional embedding, where $\alpha \in \mathbb{R}^d$. The word embedding of $Positive$, $Neutral$, or $Negative$ is used to represent the text polarity embedding $m \in \mathbb{R}^d$, $m \in \{ Positive, Negative, Neutral/None\}$. Please note that the number of polarities/classes, denoted as $m$, may vary for different datasets. For a given text $x$, our goal is to analyze the text $x$ with a model to detect its associated sentiment polarity/class  $m \in \{ Positive, Negative, Neutral/None\}$ with high accuracy. For example, for a given text \textit{"ABCAir although hour delay every single staff member ticket desk admiral club sweet pie"}, what would be the polarity/class of the text? The model needs to predict the class as either $Positive$, $Negative$, or $Neutral/None$ with high accuracy. Moreover, we are motivated to tackle the question of \textbf{how Transformer and RNN-based hybrid model enhance sentiment analysis performance}. In response, we introduce a hybrid model, RoBERTa-BiLSTM, aimed at enhancing sentiment analysis performance.

\section{Proposed RoBERTa-BiLSTM Approach} \label{proposedapproach}


In this section, we describe the proposed hybrid model, RoBERTa-BiLSTM, for sentiment analysis. The RoBERTa-BiLSTM model blends the strengths of Transformer and RNN architectures to enhance efficacy and accuracy in sentiment analysis tasks. The architecture of the proposed model is depicted in Figure \ref{proposed_architecture}. The pretrained RoBERTa model acts as the encoder in the proposed hybrid model, tokenizing all input text and efficiently mapping tokens into meaningful word embedding representations. These word embeddings, generated by the pretrained RoBERTa model, are then fed into a BiLSTM layer to capture long-range dependencies within the sequence of word embeddings. A dropout layer is inserted between RoBERTa and BiLSTM to promote model generalization and prevent overfitting. Subsequently, a dense layer is added to understand the correlation between the output of BiLSTM and sentiment class labels. Finally, the classification layer employs the Softmax function to estimate probability distributions of the classes. The overall steps of the proposed RoBERTa-BiLSTM hybrid model are outlined as pseudocode in Algorithm \ref{algo_sentiment_analysis}. The components of the proposed hybrid model are detailed below.

\begin{algorithm}
    \caption{The overall steps of the proposed sentiment analysis approach based on the RoBERTa-BiLSTM model}\label{algo_sentiment_analysis}
    \begin{algorithmic}[1]
        \STATE  \textbf{Input:} Text/Comments $X=\{x_1, x_2, x_3, \cdots, x_n\}$
        \STATE  \textbf{Output:}  Predict the underlying sentiment (e.g., \texttt{positive}, \texttt{neutral}, or \texttt{negative}) based on the given text/comments.
        
        \STATE \textbf{Lowercasing} the text to ensure uniformity.
        \STATE Set of elements for removal $E$=\{ \texttt{special ~symbols}, \texttt{URLs}, \texttt{hashtags}, \texttt{punctuation}, \texttt{special characters}, \texttt{numbers}, \texttt{the}, \texttt{an}, \texttt{a}\}
        
         
        \FOR{each text $ x \in X$} 
            \STATE {Initialize a list $W[]$ for words}
            \FOR{each word $\alpha \in x $}{
            
                \IF{$\alpha \in E $}
                    \STATE Remove the word $\alpha$ from the text $x$
                \ELSE
                    \STATE {Assign words $W [] \leftarrow \alpha$}
                \ENDIF
                     
                }
            \ENDFOR

    \ENDFOR 

    \STATE \textbf{Lemmatization} is applied to $W$.
    
    \STATE \textbf{Perform RoBERTa Tokenization} to assign a unique \texttt{Input ID}, \texttt{Token ID}, and \texttt{Attention Mask} to each token.
    
    \STATE \textbf{RoBERTa Framework} comprises 12 layers, each composed of 768 hidden states.It is utilized to generate word embeddings,  $E \in \mathbb{R} ^{l \times d}$, as per Eq (\ref{embedding_matrix}). 

    \STATE \textbf{Dropout Layer} is implemented to promote model generalization and mitigate overfitting.

    \STATE \textbf{BiLSTM Layer} is added to extract features and enhance the model's predictive accuracy by leveraging both contextual information from RoBERTa and long-range dependencies between tokens.
    
    \STATE \textbf{Flatten Layer} is added to reshape the input from multi-dimensional tensor to one-dimensional tensor for the subsequent dense layer.
    
    \STATE \textbf{Dense Layer} is added to connect all inputs from the preceding layer to all activation units in the subsequent layer.
    
    \STATE \textbf{Classification Layer} is utilized to predict the sentiment of the text/comment.
    \end{algorithmic}
\end{algorithm}

\subsection{RoBERTa}




RoBERTa, an extension of the BERT model, emerges as a powerful asset in the realm of NLP, engineered to enhance the effectiveness of natural language understanding (NLU) tasks. With its 12-layer architecture and 768 hidden states per layer, RoBERTa aims to surpass the limitations of its predecessor by amalgamating extensive pretraining with fine-tuning strategies. First introduced by Facebook AI researchers in 2019 \cite{liu2019roberta}, RoBERTa embodies the Robustly Optimized BERT approach, striving to enhance performance across a spectrum of NLU tasks including text classification, question answering, and natural language inference. Both BERT and RoBERTa employ the Transformer architecture to facilitate sequence-to-sequence tasks, relying on self-attention mechanisms to discern dependencies between inputs and outputs. RoBERTa adopts a self-supervised approach, where it undergoes pretraining on raw text data without requiring human annotations. It autonomously generates inputs and corresponding labels from the provided texts. RoBERTa's training data comprises a vast corpus, totaling ten times ($10 \times$) the size of BERT's training data. This extensive dataset comprises four primary sources: ($i$) BookCorpus combined with English Wikipedia (16 GB) \cite{zhu2015aligning}, $ii$) OpenWebText (38 GB) \cite{gokaslan2019openwebtext},($iii$) CC-News (76 GB), ( and ($iv$) Stories (31 GB). Together, these sources contribute to an impressive 161 GB of text data.

The proposed hybrid model builds upon RoBERTa as its foundational layer, harnessing its advanced capabilities in contextual learning and tokenization. Unlike BERT's static masking, RoBERTa adopts dynamic masking, enabling the model to extract insights from diverse input sequences. Additionally, RoBERTa's training on a large text corpus and utilization of byte-level Byte Pair Encoding for tokenization enhance its computational efficiency and vocabulary robustness. The proposed model utilizes the pretrained RoBERTa tokenizer to break down raw text into subword tokens, preserving semantic meaning while mitigating the impact of out-of-vocabulary words. Each token is then assigned a unique input ID, token ID, and attention mask, facilitating focused processing within the RoBERTa framework. The harmonious integration of RoBERTa's capabilities with additional refinements highlights its crucial role in advancing NLP performance and comprehension across diverse domains, pushing the boundaries of language understanding to unprecedented heights.




\subsection{Dropout Layer}

The dropout layer emerges as a pivotal technique in DL and Transformer models, playing a crucial role in preventing overfitting and improving model generalization. Originating from the seminal work of Srivastava et al. \cite{srivastava2014dropout}, dropout involves randomly deactivating a fraction of neurons within a layer during each training iteration, thereby reducing the interdependence between neurons and preventing the network from memorizing noise in the training data. This regularization technique has been widely adopted in various neural network architectures, including CNNs, RNNs, and Transformers, with significant success in enhancing performance on tasks such as image classification, NLP, and speech recognition \cite{goodfellow2016deep}. The versatility and effectiveness of dropout make it a fundamental component in the toolkit of DL practitioners, facilitating the training of more robust and generalizable models.  The dropout layer is placed between RoBERTa and BiLSTM layers in the proposed hybrid model.

\subsection{BiLSTM}


BiLSTM, short for Bidirectional Long Short-Term Memory, represents a type of RNN architecture comprising two LSTM networks. One LSTM processes the input sequence from left to right (known as the forward LSTM), while the other processes it from right to left (known as the backward LSTM) \cite{graves2008offline, seabe2023forecasting}. This bidirectional processing nature allows BiLSTM to comprehend sequences more effectively, rendering it particularly valuable in NLP tasks such as sentiment analysis \cite{vimali2021text, bhuvaneshwari2022sentiment, kota2022high}, entity recognition \cite{yang2020residual, lin2017multi}, and machine translation. Hochreiter and Schmidhuber \cite{hochreiter1997long} introduced the LSTM architecture to tackle the vanishing gradient problem in RNNs. Graves and Schmidhuber \cite{graves2005framewise, graves2008offline} further investigated the effectiveness of BiLSTMs in tasks such as phoneme classification and handwriting recognition. Their work demonstrated the superiority of BiLSTMs over unidirectional models in capturing rich contextual information. Consequently, BiLSTMs have emerged as a widely adopted architecture for various sequential data analysis tasks, offering enhanced performance and robustness.  In the proposed hybrid model, the output of the RoBERTa layer is passed through a dropout layer before being fed into the BiLSTM layer. The BiLSTM possesses the ability to retain past information and effectively manage long-range dependencies within the input. Therefore, integrating a BiLSTM as a feature extractor enhances the model's predictive accuracy by leveraging both the contextual information from RoBERTa and the long-range dependencies between tokens. The BiLSTM model architecture is described by the following set of equations \cite{graves2005framewise, rahman2023multilingual}.

\subsubsection{Input Gate ($i_t$)}

\begin{align}
i_t &= \sigma(W_{ix}^f x_t + W_{ih}^f h_{t-1}^f + W_{ic}^f c_{t-1}^f + b_i^f) \nonumber \\
&\quad \odot \sigma(W_{ix}^b x_t + W_{ih}^b h_{t+1}^b + W_{ic}^b c_{t+1}^b + b_i^b) \label{eq:bilstm_input_gate}
\end{align}

Equation \eqref{eq:bilstm_input_gate} controls the flow of information into the cell state $C_t$ at time step $t$ in both the forward and backward directions. It combines the input from the forward LSTM ($W_{ix}^f x_t + W_{ih}^f h_{t-1}^f + W_{ic}^f c_{t-1}^f + b_i^f$) and the backward LSTM ($W_{ix}^b x_t + W_{ih}^b h_{t+1}^b + W_{ic}^b c_{t+1}^b + b_i^b$) using element-wise multiplication. The sigmoid function $\sigma$ squashes the input to a range between 0 and 1, determining the extent to which each component affects the input gate.

\subsubsection{Forget Gate ($f_t$)}

\begin{align}
f_t &= \sigma(W_{fx}^f x_t + W_{fh}^f h_{t-1}^f + W_{fc}^f c_{t-1}^f + b_f^f) \nonumber \\
&\quad \odot \sigma(W_{fx}^b x_t + W_{fh}^b h_{t+1}^b + W_{fc}^b c_{t+1}^b + b_f^b) \label{eq:bilstm_forget_gate}
\end{align}

Equation \eqref{eq:bilstm_forget_gate} determines which information from the previous cell state $c_{t-1}$ should be discarded or forgotten in both the forward and backward directions. It combines the forget gate computations from the forward and backward LSTMs using element-wise multiplication.

\subsubsection{Cell State Update ($C_t$)}

\begin{align}
C_t &= f_t \odot C_{t-1} + i_t \odot \tanh(W_{cx}^f x_t + W_{ch}^f h_{t-1}^f + b_c^f) \nonumber \\
&\quad + i_t \odot \tanh(W_{cx}^b x_t + W_{ch}^b h_{t+1}^b + b_c^b) \label{eq:bilstm_cell_state}
\end{align}

Equation \eqref{eq:bilstm_cell_state} updates the cell state $C_t$ by combining information from both the forward and backward directions. It combines the contributions from the input gate in both directions and the tanh activations of the candidate values.

\subsubsection{Output Gate ($o_t$)}

\begin{align}
o_t &= \sigma(W_{ox}^f x_t + W_{oh}^f h_{t-1}^f + W_{oc}^f c_{t}^f + b_o^f) \nonumber \\
&\quad \odot \sigma(W_{ox}^b x_t + W_{oh}^b h_{t+1}^b + W_{oc}^b c_{t}^b + b_o^b) \label{eq:bilstm_output_gate}
\end{align}

Equation \eqref{eq:bilstm_output_gate} regulates which parts of the cell state $C_t$ should be used to compute the hidden state $h_t$ at the current time step $t$ in both the forward and backward directions. It combines the output gate computations from both directions using element-wise multiplication.

\subsubsection{Hidden State ($h_t$)}

\begin{align}
h_t &= o_t \odot \tanh(C_t) \label{eq:bilstm_hidden_state}
\end{align}

Equation \eqref{eq:bilstm_hidden_state} computes the hidden state $h_t$ by applying the output gate to the hyperbolic tangent of the cell state $C_t$.

\subsubsection{Output ($y_t$)}

\begin{align}
y_t &= \text{Softmax}(W_{hy} h_t + b_y) \label{eq:bilstm_output}
\end{align}

Equation \eqref{eq:bilstm_output} generates the output prediction at time step $t$ by applying a Softmax function to the linear transformation of the hidden state $h_t$.


\begin{figure*}
\captionsetup{justification=centering}
     \centering
     \begin{subfigure}[b]{0.32\textwidth}
 \centering
  \includegraphics[width=.85\linewidth]{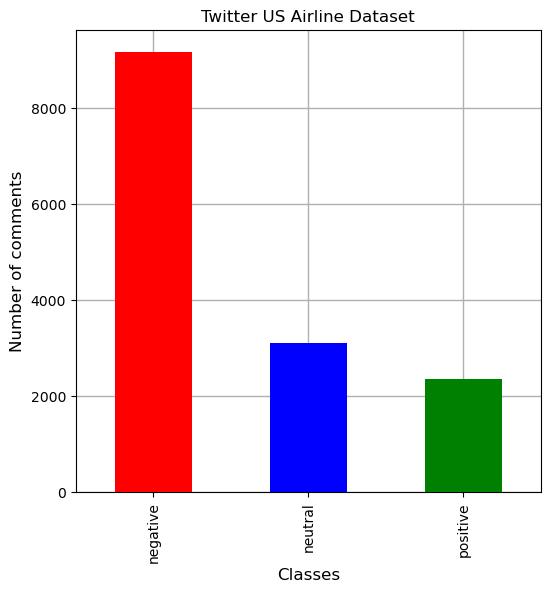}
  \caption{Twitter US Airline}
   \label{twitter_us_airline}
     \end{subfigure}
     \hfill
       \begin{subfigure}[b]{0.32\textwidth}
     \centering
  \includegraphics[width=.85\linewidth]{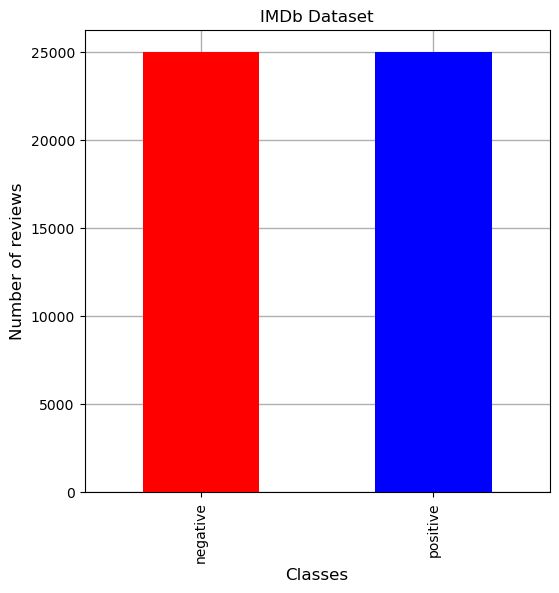}
  \caption{IMDb Review}
   \label{imdb_review}
     \end{subfigure}
     \hfill
    \begin{subfigure}[b]{0.32\textwidth}
     \centering
  \includegraphics[width=.85\linewidth]{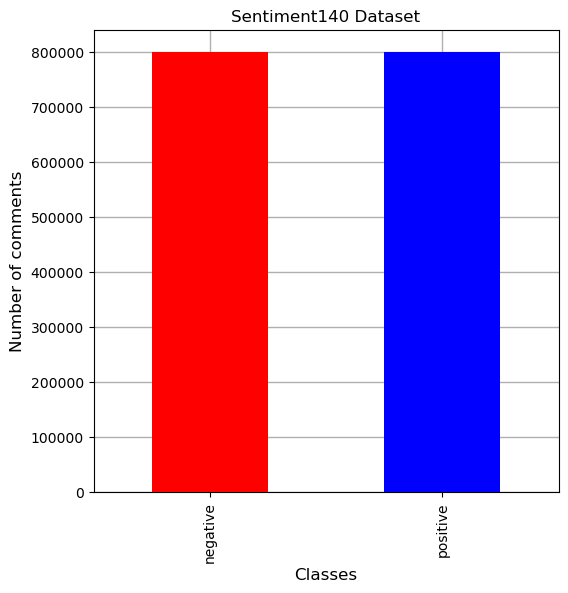}
  \caption{Sentiment140 }
   \label{sentiment_140}
     \end{subfigure}
     \hfill

       \caption{The sample distribution of the Twitter US Airline,  IMDb Review, and Sentiment140 datasets}
        \label{main_datasets}
\end{figure*}

\subsection{Flatten Layer}

In the proposed architecture of the RoBERTa-BiLSTM model, the flatten layer plays a pivotal role in facilitating the transition from the output of the BiLSTM layer to the subsequent dense layer. Specifically, the flatten layer serves to reshape the output of the BiLSTM layer into a format compatible with the input requirements of the dense layer. The output of the BiLSTM layer typically comprises a tensor with multiple dimensions, representing the sequential nature of the input data processed by the bidirectional LSTM units. However, the subsequent dense layer expects its input to be in the form of a flat tensor with a one-dimensional shape. To bridge this disparity in tensor shapes, the flatten layer performs the essential operation of transforming the multi-dimensional tensor output of the BiLSTM layer into a one-dimensional tensor \cite{tan2023roberta}. This transformation is achieved by unrolling all the elements of the tensor, effectively converting it into a linear sequence.
   


\subsection{Dense Layer}

The dense layer, also known as the fully connected layer, plays a crucial role in capturing the relationships between the hidden states generated by the BiLSTM layer and the class labels. This layer establishes dense connectivity by connecting all neurons from the preceding layer to those in the subsequent layer. In the proposed model, two dense layers are incorporated. The first dense layer encodes the relationship between the flattened output of the BiLSTM and the class labels. It captures the underlying patterns in the data to facilitate classification. The second dense layer performs the final classification by generating a probability distribution using the Softmax activation function for the classes. The Softmax function squashes the output values to fall within the range of [0, 1], ensuring that the sum of the probabilities equals 1.




\subsection{Softmax Layer}

The Softmax layer serves as the top layer in the proposed hybrid model for sentiment classification. Also referred to as the classification or output layer, it applies the Softmax function to generate probability distributions for the classes. This function can be described as follows:

 \begin{equation}
    Softmax(\emph{\textbf{O}})_j=\frac{e^{\emph{\textbf{O}}_j}}{\sum_{k=1}^M e^{\emph{\textbf{O}}_k}}
    \label{softmaxeq}
\end{equation}

$M$ denotes the number of sentiment classes. The numerator, $e^{\emph{\textbf{O}}_j}$, represents the exponential function applied to each element of $\emph{\textbf{O}}$. The denominator, $\sum_{k=1}^M e^{\emph{\textbf{O}}_k}$, denotes the sum of the exponential functions of all elements.

\section{Dataset} \label{dataset}
 In this study, we utilized three publicly available datasets: IMDb, Twitter US Airline, and Sentiment140, to evaluate the performance of the proposed RoBERTa-BiLSTM model. Figure \ref{main_datasets} illustrates the sample data distributions within these three datasets. The Twitter US Airline dataset \cite{tan2022roberta} consists of 14,640 tweets categorized into three sentiment classes: positive, neutral, and negative. These tweets were gathered from customers of six major American Airlines, namely Delta, US-Airways, Virgin America, Southwest, and United, for sentiment analysis. The distribution of tweets across sentiment classes reveals an imbalance, with 62.69\% of tweets classified as negative, 16.14\% as positive, and 21.17\% as neutral. 
 The IMDb dataset \cite{maas2011learning} comprises a total of 50,000 reviews, evenly split between positive and negative reviews, resulting in a balanced dataset with 50\% of samples allocated to each class. The Sentiment140 dataset \cite{go2009twitter} is a sizable collection of approximately 1.6 million tweets designed for sentiment analysis. The Sentiment140 dataset was compiled from Twitter by Stanford University in 2009. This dataset features an equal distribution of tweets across positive and negative classes, with each class representing 50\% of the dataset.

\subsection{Data Preprocessing}

\begin{figure} [h]
\centering
  \includegraphics[width=\linewidth]{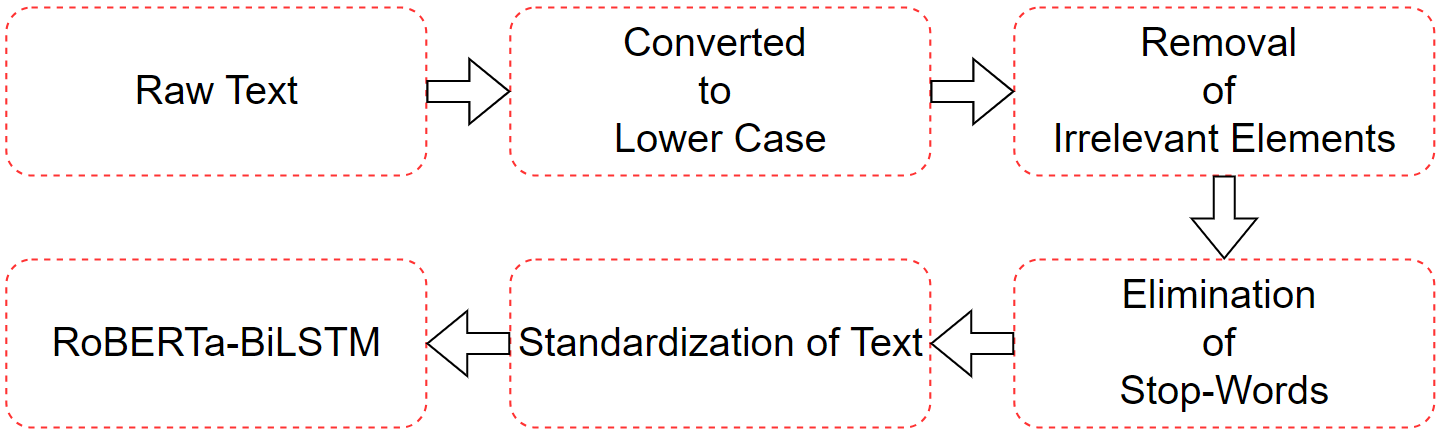}
  \caption{The data preprocessing steps of the proposed approach}
  \label{data_preprocessing}
\end{figure}

In sentiment analysis, data preprocessing plays a crucial role in filtering out irrelevant elements that could potentially hinder the performance of the ML models. Figure \ref{data_preprocessing} illustrates the comprehensive data preprocessing steps involved in the proposed RoBERTa-BiLSTM approach. Since the texts or comments within datasets are collected from users of platforms like IMDb or Twitter, they may contain a mix of upper and lower-case text. Therefore, case folding is an essential preprocessing step to ensure a consistent text case. All texts are converted to lowercase to maintain uniformity. Moreover, irrelevant elements such as special symbols, punctuation, hashtags, URLs, special characters, and numbers are removed from the text. Another significant aspect of data preprocessing involves the elimination of stop-words, which carry little meaning in sentiment analysis. Stop-words, such as \textit{the}, \textit{an}, and \textit{a}, are frequently occurring words in the text that are syntactically important but semantically less relevant. Furthermore, lemmatization \cite{balakrishnan2014stemming} is applied to bring words to their original form, enhancing semantic consistency. For instance, the word \textit{caring} is transformed into \textit{care} after a lemmatization operation. Although both lemmatization and stemming serve the same purpose of reducing words to their base forms, lemmatization offers several advantages over stemming. The data preprocessing steps standardize the raw data by removing irrelevant information. This allows the LLMs to effectively process the filtered data, thereby enhancing the performance of sentiment analysis.

\section{Hyperparameters} \label{hyper_tuning}

In the proposed RoBERTa-BiLSTM model, various sets of hyperparameters are employed to achieve superior results. The selection of an optimal set of hyperparameters is pivotal for accurately analyzing sentiments. RoBERTa and BERT, both LLMs, are separately combined with different RNN architectures, including LSTM, GRU, and BiLSTM. During the training process, different learning rates ($l$) (i.e., $l=\{0.0001, 0.00001, 0.000001\}$) and hidden units ($h$) (i.e., $h=\{128, 256, 512\}$) of the RNN are utilized. Please note that the number of hidden units ($h$) will be doubled ($2 \times h$) (i.e., $\overrightarrow {h}$ + $\overleftarrow{h})$ for the BiLSTM due to its forward ($\rightarrow$) and backward ($\leftarrow$) data processing nature. The loss function plays a crucial role in network optimization by computing the overall model loss in each training epoch. Given that sentiment analysis involves a multi-class classification problem, categorical cross-entropy is chosen as the loss function ($\mathbf{L}$). It is defined as follows:
\begin{equation} \label{lossfunction}
     \mathbf{L}(p)=-\sum_{i=1}^M y_i ~log(\hat{y_i})
 \end{equation}
 
 where $p$ denotes the model parameter; $M$ denotes the number of classes; and $y_i$ and $\hat{y_i}$ are the true and predicted labels, respectively for the $i^{th}$ sample. Table \ref{hyperpameters} provides a summary of the hyperparameters and their corresponding values used in the experiments.

\begin{table}[!htbp]
\caption{The sets of hyperparameters for model fine-tuning}
\begin{center}
\begin{tabular}{p{3.2cm}|p{4.5cm}}
\hline \hline
\textbf{Hyperparameters} & \textbf{Values} \\
\hline \hline
Large Language Models (LLMs) & BERT (\texttt{bert-base-uncased} \cite{devlin2018bert}), RoBERTa (\texttt{roberta-base} \cite{liu2019roberta})  \\ \hline
Recurrent Neural Networks (RNNs) & GRU, LSTM, BiLSTM  \\ \hline
Optimization method & AdamW, SGD, RMSprop, Rprop \\ \hline
Loss function ($\mathbf{L}$)  & Categorical Cross Entropy (\texttt{cross\_entropy}) \\ \hline

Epochs ($epoch$)& 5 \\ \hline
Dropout ($d$)& 0.1 \\ \hline
Learning rates ($l$) & 0.0001, 0.00001, 0.000001\\ \hline
Hidden units ($h$) of RNNs & 128, 256, 512\\ \hline
Datasets & IMDb, Twitter US Airline, Sentiment140\\ \hline
Training data & 90\%\\ \hline
Validation data& 5\%\\ \hline
Testing data& 5\%\\ \hline

\end{tabular}
\label{hyperpameters}
\end{center}
\end{table}

\section{Experimental Results} \label{experimental_results}
In this section, we present the experimental results of the RBERTa-base, RoBERTa-GRU,  RoBERTa-LSTM, RoBERTa-BiLSTM, BERT-GRU,  BERT-LSTM, and BERT-BiLSTM models on the IMDb, Twitter US Airline, and Sentiment140 datasets. Furthermore, the hyperparameters are meticulously fine-tuned to determine the most optimal settings and parameters for each model. Finally, we compare the overall performance of the proposed RoBERTa-BiLSTM model against state-of-the-art models.

\subsection{Implementation Details}

The experiments are conducted within the operating system environment of Ubuntu 22.04.4 LTS 64-bit. The hardware specifications are detailed as follows: Processor: AMD Ryzen 9 3950X 16-core processor with 32 threads, RAM: 64GB, Graphics: NVIDIA GeForce RTX 3090/PCIe/SSE2, Graphics Memory: 24GB, and Disk Capacity: 500GB.

\begin{table*}[!t]
\caption{Quantitative results ($\mathbf{F1}_w$, $\mathbf{P}_w$, and $\mathbf{R}_w$) for sentiment analysis using the RoBERTa-base model are based on the IMDb, Twitter US Airline, and Sentiment140 datasets} \label{roberta_base_model_p_r_f1}
\centering
\begin{tabular}{p{1cm}|p{1.4cm}||p{.9cm}|p{.9cm}|p{.9cm} ||p{.9cm}|p{.9cm}|p{.9cm}||p{.9cm}|p{.9cm}|p{.9cm}}
\hline \hline

\multirow{2}{*}{\textbf{\makecell{Learning \\Rate ($l$)}}} & \multirow{2}{*}{\textbf{\makecell{Model \\Evaluation}}}  & \multicolumn{3}{c||}{IMDb Dataset}  & \multicolumn{3}{c||}{Twitter US Airline Dataset} & \multicolumn{3}{c}{Sentiment140 Dataset} \\ \cline{3-11}

&  & \centering $\mathbf{F1}_w$ & \centering $\mathbf{P}_w$ &  \makecell{$\mathbf{R}_w$}  & \centering $\mathbf{F1}_w$ & \centering $\mathbf{P}_w$ & \makecell{$\mathbf{R}_w$}  & \centering $\mathbf{F1}_w$ & \centering $\mathbf{P}_w$ &  \makecell{$\mathbf{R}_w$} \\ \hline \hline

\multirow{3}{*}{0.0001} & Training &  0.333259 &	0.249933 &	0.499933 &	0.679082 &	0.677392 &	0.728218 &	0.333219 &	0.249897 &	0.499897\\ \cline{2-11}
& Validation & 	0.342260 &	0.258064 & 	0.508000 & 	0.680925 &	0.677001 	& 0.733607 & 	0.337725 &	0.253960 & 	0.503944 \\ \cline{2-11}
& Test & 	0.325805 &	0.243246 &	0.493200 &	0.675668 &	0.690883 &	0.718579 &	0.331457 &	0.248312 &	0.498310 \\ \hline

\multirow{3}{*}{0.00001} & Training &  0.956633	& 0.957143	& 0.956644 &	0.911728 &	0.912569	& 0.911278 & 0.859462	& 0.859888 &	0.859499\\ \cline{2-11}
& Validation & 	0.908376 &	0.909582 &	0.908400 & 0.780347 &	0.784632 &	0.777322 & 0.818549 &	0.818888 &	0.818617 \\ \cline{2-11}
& Test & 	\textbf{0.913101}	& \textbf{0.914400}	& \textbf{0.913200} &	\textbf{0.801151} &	\textbf{0.807024} &	\textbf{0.797814} & \textbf{0.821697} &	\textbf{0.822092} & \textbf{0.821734} \\ \hline

\multirow{3}{*}{0.000001} & Training &  0.917354 & 	0.917387 &	0.917356 &	0.748574 	& 0.746235 &	0.752353 & 	0.333219 &	0.249897 & 	0.499897 \\ \cline{2-11}
& Validation & 	0.903604 &	0.903632 &	0.903600  &	0.727966 &	0.727360 &	0.730874 &	0.337725 &	0.253960 &	0.503944\\ \cline{2-11}
& Test & 	0.900390 &	0.900429 &	0.900400 & 	0.739893 & 0.738560 	& 0.745902 & 	0.331457 &	0.248312 &	0.498310 \\ \hline

\end{tabular}
\end{table*}

 \begin{table*}[!t]
\caption{Quantitative results ($\mathbf{F1}_w$, $\mathbf{P}_w$, and $\mathbf{R}_w$)  for sentiment analysis using the RoBERTa-GRU model are based on the IMDb, Twitter US Airline, and Sentiment140 datasets} \label{roberta_gru_model_p_r_f1}
\centering
\begin{tabular}{p{1cm}|p{1.4cm}|p{1.1cm}||p{.9cm}|p{.9cm}|p{.9cm} ||p{.9cm}|p{.9cm}|p{.9cm}||p{.9cm}|p{.9cm}|p{.9cm}}
\hline \hline

\multirow{2}{*}{\textbf{\makecell{Learning \\Rate ($l$)}}} & \multirow{2}{*}{\textbf{\makecell{Model \\Evaluation}}} & \multirow{2}{*}{\textbf{\makecell{Hidden \\Units ($h$)}}} & \multicolumn{3}{c||}{IMDb Dataset}  & \multicolumn{3}{c||}{Twitter US Airline Dataset} & \multicolumn{3}{c}{Sentiment140 Dataset} \\ \cline{4-12}

 &  &  & \centering $\mathbf{F1}_w$ & \centering $\mathbf{P}_w$ &  \makecell{$\mathbf{R}_w$}  & \centering $\mathbf{F1}_w$ & \centering $\mathbf{P}_w$ & \makecell{$\mathbf{R}_w$}  & \centering $\mathbf{F1}_w$ & \centering $\mathbf{P}_w$ &  \makecell{$\mathbf{R}_w$} \\ \hline \hline

\multirow{9}{*}{0.0001} & \multirow{3}{*}{Training} & 128 &	0.333407 &	0.250067	& 0.500067 &	0.633524	& 0.604036	& 0.675774 &	0.333219 &	0.249897	 & 0.499897 \\ 
& & 256 & 0.333407	& 0.250067	& 0.500067 &	0.497897	 & 0.533193	& 0.633728 &	0.333219 &	0.249897	 & 0.499897 \\ 
& & 512 &	0.333259	& 0.249933	& 0.499933  &	0.605418	& 0.541333 &	0.692927 &	0.333219 &	0.249897	 & 0.499897\\ \cline{2-12} 

& \multirow{3}{*}{Validation} & 128 & 0.324483	& 0.242064 &	0.492000 &	0.598217 & 	0.574599	& 0.633880 &  0.337725 & 	0.253960 &	0.503944\\ 
& & 256 &	0.324483	& 0.242064	& 0.492000 &	0.516451 & 	0.552385 &	0.648907 &  0.337725 & 	0.253960 &	0.503944 \\ 
& & 512 & 0.342260	& 0.258064	& 0.508000 &	0.643967 &	0.585739 &	0.722678 &  0.337725 & 	0.253960 &	0.503944\\ \cline{2-12} 
  
& \multirow{3}{*}{Test} & 128 &	0.340916 &	0.256846 &	0.506800 & 0.585397	& 0.552140	& 0.631148 & 0.331457	& 0.248312	& 0.498310 \\ 
& & 256 & 0.340916 &	0.256846	& 0.506800 & 0.477446	& 0.524948 & 	0.612022 & 0.331457	& 0.248312	& 0.498310\\ 
& & 512 & 0.325805	& 0.243246	& 0.493200 & 0.579547 & 0.516748	& 0.669399 & 0.331457	& 0.248312	& 0.498310\\ \hline

\multirow{9}{*}{0.00001} & \multirow{3}{*}{Training} & 128  &	0.955797 &	0.955939 	& 0.955800 &	0.870027 &	0.869620 &	0.873103 &	0.863399 &	0.863412 &	0.863400\\ 
& & 256 &	0.938483 &	0.938665 &	0.938489 &  0.820532	& 0.828654	& 0.832271 & 	0.863389 	& 0.863402 	& 0.863390 \\ 
& & 512 &	0.939976 &	0.940030 &	0.939978 &	0.829548 &	0.832554 &	0.838039 &	0.863479 &	0.863479 &	0.863479 \\ \cline{2-12} 

& \multirow{3}{*}{Validation} & 128 &	0.917603 &	0.917945 &	0.9176 &	0.789664 &	0.787591 	& 0.795082 &	0.819446 &	0.819461 &	0.819443\\ 
& & 256  &	0.911955 &	0.912378 &	0.912000 & 0.764612	& 0.766811	& 0.782787 &	0.819859 &	0.819892 &	0.819856 \\ 
& & 512 &	0.918005 &	0.918166 &	0.918000 &	0.782206 &	0.782099	& 0.795082 &	0.819120 &	0.819128 &	0.819117 \\ \cline{2-12} 
  
& \multirow{3}{*}{Test} & 128 &	\textbf{0.925966} &	\textbf{0.926384} &	\textbf{0.926000} &	0.793064 &	0.792070 	& 0.795082 &	0.822510 &	0.822512 &	0.822511 \\ 
& & 256 &	0.916803 &	0.916975 &	0.916800 & 	0.771077	& 0.776701	& 0.784153 &	\textbf{0.823194} &	\textbf{0.823216} &	\textbf{0.823199} \\ 
& & 512 &	0.919576 &	0.919801 &	0.919600 &	\textbf{0.790581} &	\textbf{0.794184} &	\textbf{0.799180} &	0.822123 &	0.822124 &	0.822123 \\ \hline

\multirow{9}{*}{0.000001} & \multirow{3}{*}{Training} & 128  & 0.920795	& 0.920900 & 	0.920800 &	0.808283 & 	0.806835 &	0.811020 &	0.822880	& 0.822998	& 0.822894\\ 
& & 256& 0.926133	& 0.926140	& 0.926133  &	0.813906 &	0.812793 &	0.817168 & 0.823222 &	0.823254 &	0.823226  \\ 
& & 512  & 0.926422	& 0.926423	& 0.926422 & 0.809176	& 0.807872 &	0.812234 & 0.823492	& 0.823514	& 0.823495 \\ \cline{2-12} 

& \multirow{3}{*}{Validation} & 128  & 0.904380	& 0.904489	& 0.904400 & 0.751607 &	0.751851	& 0.751366 & 0.812731 &	0.812763 &	0.812744\\ 
& & 256 & 0.914787	& 0.914858	& 0.914800 &	0.758944 &	0.755749 &	0.763661 & 0.813230	& 0.813231	& 0.813232  \\ 
& & 512  & 0.912003	& 0.912016	& 0.912000 &	0.756455 &	0.754155	& 0.759563 & 0.813305	& 0.813306	& 0.813308\\ \cline{2-12} 
  
& \multirow{3}{*}{Test} & 128  & 	0.904404	& 0.904513	& 0.904400 & 0.773350	& 0.772348 &	0.774590 & 0.816133	& 0.816201	& 0.816137\\ 
& & 256 & 	0.909602	& 0.909615	& 0.909600 & 0.772407 &	0.770946 &	0.774590 & 0.816200	& 0.816222	& 0.816200 \\ 
& & 512  & 0.911598	& 0.911599	& 0.911600  & 	0.775682	& 0.774644 &	0.777322 & 0.816262	& 0.816284	& 0.816263\\ \hline 

\end{tabular}
\end{table*}

\subsection{Evaluation Metrics}

In the context of sentiment analysis, it is important to comprehend the underlying meaning of comments/texts and then classify them as positive, negative, or neutral. The proposed hybrid RoBERTa-BiLSTM model serves as a classifier designed specifically for sentiment analysis. To assess the performance of the classifier, a confusion matrix is employed. Metrics such as accuracy, precision, recall, and F1-score are derived using elements such as TP, FP, TN, and FN from the confusion matrix \cite{rahman2023multilingual}. Accuracy ($\mathbf{A}$), for instance, is defined as the ratio of correct predictions to the total number of predictions, as follows:

 \begin{equation}\label{class_accuracy}
 \mathbf{A}=\frac{1}{N}\sum^{|M|}_{i=1} \sum_{x:f(x)=i} \Upsilon(f(x) = \hat{f}(x))
 \end{equation}
 
where $\Upsilon$ is a function that returns 1 if the class is true and 0, otherwise. Here, $f(x) \in M =\{1, 2, 3, \cdots \}$. We also computed the precision, recall, and F1-score under weighted-average settings. The F1-score is derived from the mean of all F1-scores of individual classes, taking into account the support of each class. The term \texttt{support} refers to the number of instances per class, while \texttt{weight} denotes the ratio of instances of each class to the total instances. The weighted-precision ($\mathbf{P}_w$), recall ($\mathbf{R}_w$), and F1-score ($\mathbf{F1}_w$) are calculated in Eqs. (\ref{weight_p})-(\ref{weight_f1}):

   \begin{equation}\label{weight_p}
     \mathbf{P}_w= \frac{1}{|Q|} \sum_{i=1}^{|
     M|} \frac{TP_i}{TP_i + FP_i} \times |q_i| = \frac{\sum_{i=1}^{|M|} \mathbf{P}_i \times |q_i|}{|Q|}
 \end{equation}
 
   \begin{equation}\label{weight_r}
     \mathbf{R}_w= \frac{1}{|Q|} \sum_{i=1}^{|M|} \frac{TP_i}{TP_i + FN_i} \times |q_i| = \frac{\sum_{i=1}^{|M|} \mathbf{R}_i \times |q_i|}{|Q|}
 \end{equation}

   \begin{equation}\label{weight_f1}
     \mathbf{F1}_w= \frac{1}{|Q|} \sum_{i=1}^{|M|} \mathbf{F1}_i \times |q_i|
 \end{equation}

where $|Q|$ is the sum of all supports and $|q_i|$ is the support of the $i$ class.

\begin{figure} [h]
     \centering
    
         \captionsetup{justification=centering}
         \begin{tikzpicture}[scale=1]
            \begin{axis}[
    ybar=.13cm,
    every node near coord/.append style={font=\tiny},
    legend style={font=\tiny},
    tick label style={font=\tiny},
    ylabel near ticks, ylabel shift={-6pt},
    width=\linewidth,
    height=5.2cm,
    every node near coord/.append style={
                        anchor=west,
                        rotate=75
                },
    enlargelimits=.40,
    enlarge y limits={0.1,upper},
    legend style={at={(0.5,-0.22)},
    anchor=north, legend columns=-1},
    ymin=30, 
    ylabel={$\mathbf{A}$ (\%)},
    symbolic x coords={IMDb, Twitter US Airline, Sentiment140},
    xtick=data,
    ytick={10, 20, 30, 40, 50, 60, 70, 80, 90, 100},
    grid=both,
    nodes near coords,
    nodes near coords align={vertical},
    bar width=12pt,
    label style={font=\footnotesize},
    ]
\addplot [draw=red, semithick, pattern=crosshatch, pattern color = red] coordinates {(IMDb,49.32) (Twitter US Airline,71.86) (Sentiment140,49.83)};  

\addplot [draw=blue, semithick, pattern=north east lines,  pattern color = blue] coordinates {(IMDb,91.32) (Twitter US Airline,79.78) (Sentiment140,82.17)};  

\addplot [draw=teal, semithick, pattern=grid,  pattern color = teal] coordinates {(IMDb,90.04) (Twitter US Airline,74.59) (Sentiment140,49.83)}; 


\legend{$l=0.0001$, $l=0.00001$, $l=0.000001$}
\end{axis}
 
        \end{tikzpicture}     
    
          \caption{Assessing the test accuracy of the RoBERTa-base model using various hyperparameters: $l=\{0.0001, 0.00001, 0.000001\}$. The model is trained for 5 \texttt{epochs} using the \texttt{AdamW} optimizer.}
        \label{roberta_base_accuracy_imdb_twitter_sentment140}
\end{figure}
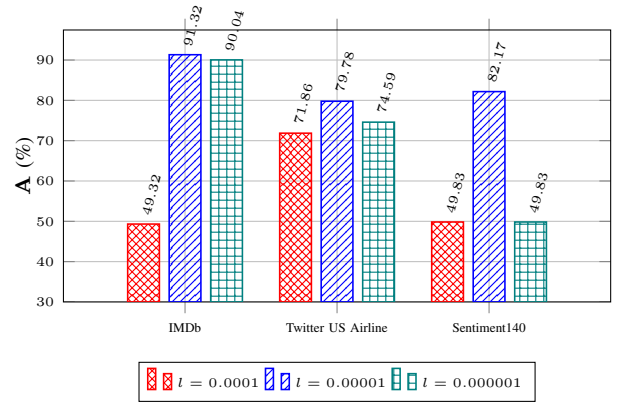

\subsection{Results}

Comprehensive experiments are conducted using three different datasets (e.g., IMDb, Twitter US Airline, and Sentiment140) and various hyperparameter settings to showcase the effectiveness of the proposed model. Table \ref{roberta_base_model_p_r_f1} presents the quantitative classification results of $\mathbf{F1}_w$, $\mathbf{P}_w$, and $\mathbf{R}_w$  for the RoBERTa-base model with learning rates of $l = 0.0001$, $0.00001$, and $0.000001$. RoBERTa-base model is a type of model that is not combined with RNNs. The model is trained for 5 epochs using the \texttt{AdamW} optimizer. It is observed that the RoBERTa-base model achieved $\mathbf{F1}_w$, $\mathbf{P}_w$, and $\mathbf{R}_w$ scores of 91.31\%, 91.44\%, and 91.32\%, respectively for the IMDb dataset; 80.12\%, 80.70\%, and 79.78\%, respectively for the Twitter US Airline dataset; and 82.17\%, 82.21\%, and 82.17\%, respectively for the Sentiment140 dataset with $l=0.00001$. Similarly, the model achieved higher $\mathbf{F1}_w$, $\mathbf{P}_w$, and $\mathbf{R}_w$ scores during training and validation. We experimented with faster learning rates ($l = 0.0001$) and slower learning rates ($l = 0.000001$) across the three datasets. However, the RoBERTa-base model failed to achieve better results compared to the outcomes obtained with $l = 0.00001$. Figure \ref{roberta_base_accuracy_imdb_twitter_sentment140} illustrates that the RoBERTa-base model achieved the comparatively higher $\mathbf{A}$ of 91.32\%, 79.78\%, and 82.17\% for the IMDb, Twitter US Airline, and Sentiment140 datasets, respectively, when using $l=0.00001$.

 \begin{table*}[!t]
\caption{Quantitative results ($\mathbf{F1}_w$, $\mathbf{P}_w$, and $\mathbf{R}_w$)  for sentiment analysis using the RoBERTa-LSTM model are based on the IMDb, Twitter US Airline, and Sentiment140 datasets} \label{roberta_lstm_model_p_r_f1}
\centering
\begin{tabular}{p{1cm}|p{1.4cm}|p{1.1cm}||p{.9cm}|p{.9cm}|p{.9cm} ||p{.9cm}|p{.9cm}|p{.9cm}||p{.9cm}|p{.9cm}|p{.9cm}}
\hline \hline

\multirow{2}{*}{\textbf{\makecell{Learning \\Rate ($l$)}}} & \multirow{2}{*}{\textbf{\makecell{Model \\Evaluation}}} & \multirow{2}{*}{\textbf{\makecell{Hidden \\Units ($h$)}}} & \multicolumn{3}{c||}{IMDb Dataset}  & \multicolumn{3}{c||}{Twitter US Airline Dataset} & \multicolumn{3}{c}{Sentiment140 Dataset} \\ \cline{4-12}

 &  &  & \centering $\mathbf{F1}_w$ & \centering $\mathbf{P}_w$ &  \makecell{$\mathbf{R}_w$}  & \centering $\mathbf{F1}_w$ & \centering $\mathbf{P}_w$ & \makecell{$\mathbf{R}_w$}  & \centering $\mathbf{F1}_w$ & \centering $\mathbf{P}_w$ &  \makecell{$\mathbf{R}_w$} \\ \hline \hline

\multirow{9}{*}{0.0001} & \multirow{3}{*}{Training} & 128 & 0.333259	& 0.249933 &	0.499933 & 0.483979  & 0.393857 & 0.627580	&	0.333219 &	0.249897	 & 0.499897 \\ 
& & 256 & 0.730056	& 0.767716	& 0.737644 & 0.606745	& 0.552990	& 0.698467 &	0.333219 &	0.249897	 & 0.499897\\ 
& & 512 &	0.333259	& 0.249933	& 0.499933 &	0.624402 &	0.657810 &	0.689891 &	0.333219 &	0.249897	 & 0.499897 \\ \cline{2-12} 

& \multirow{3}{*}{Validation} & 128 & 0.342260 &	0.258064 &	0.508000 & 0.503843	& 0.414018	& 0.643443 &  0.337725 & 	0.253960 &	0.503944\\ 
& & 256 &	0.729988	& 0.763215 &	0.736000 & 0.633612	& 0.590176	& 0.717213 &  0.337725 & 	0.253960 &	0.503944\\ 
& & 512 &   0.342260	& 0.258064	& 0.508000 & 0.626151 &	0.587101	& 0.684426 &  0.337725 & 	0.253960 &	0.503944\\ \cline{2-12} 
  
& \multirow{3}{*}{Test} & 128 & 0.325805	& 0.243246	& 0.493200	&	0.448003	& 0.358035	& 0.598361 & 0.331457	& 0.248312	& 0.498310\\ 
& & 256 & 0.721702	& 0.755312	& 0.729600  & 0.597086	& 0.555527	& 0.691257 & 0.331457	& 0.248312	& 0.498310\\ 
& & 512 &   0.325805	& 0.243246	& 0.493200 &	0.604107	& 0.555312	& 0.677596 & 0.331457	& 0.248312	& 0.498310\\ \hline

\multirow{9}{*}{0.00001} & \multirow{3}{*}{Training} & 128  & 0.953644 &	0.953644 	& 0.953644  &	0.828337 &	0.830639 &	0.826579 &	0.860572 	& 0.860574 	& 0.860572	\\ 
& & 256 &	0.938178 &	0.938186 & 	0.938178 & 0.871825	& 0.871062	& 0.873862 &	0.844569	& 0.844902	& 0.844603\\ 
& & 512 &	0.938999 &	0.939022 &	0.939000 &	0.877842	& 0.877212 &	0.879781 & 	0.862044 &	0.862074 &	0.862047 \\ \cline{2-12} 

& \multirow{3}{*}{Validation} & 128 &	0.919198 &	0.919200 &	0.919200 & 0.771294 &	0.775925 &	0.767760 &	0.820595 &	0.820595 &	0.820595\\ 
& & 256  &	0.915990 &	0.916037 &	0.916000  & 0.783467	& 0.781502	& 0.786885 & 0.818002	& 0.818426	& 0.818028\\ 
& & 512 &	0.917602 &	0.917608 &	0.917600 &	0.778456 &	0.776641 &	0.782787 &	0.819945 &	0.820022 &	0.819944 \\ \cline{2-12} 
  
& \multirow{3}{*}{Test} & 128 &	\textbf{0.920399} &	\textbf{0.920400} &	\textbf{0.920400} & 0.795464 &	0.803598 &	0.790984 &	0.822586 & 	0.822589 &	0.822586\\ 
& & 256 &	0.918003 &	0.918030 &	0.918000 & \textbf{0.803185}	& \textbf{0.804711}	& 0.\textbf{803279} &	0.821310 &	0.821626	& 0.821359\\ 
& & 512 &	0.917596 &	0.917610 &	0.917600  &	0.799630 &	0.801161 &	0.800546 &	\textbf{0.822894} &	\textbf{0.822914} &	\textbf{0.822899} \\ \hline

\multirow{9}{*}{0.000001} & \multirow{3}{*}{Training} & 128  &  0.923443 &	0.923466	& 0.923444  &	0.805770 &	0.804582 &	0.809047 & 0.822898	& 0.822926 &	0.822901\\ 
& & 256 &  0.919124 &	0.919325 &	0.919133 & 0.807996	& 0.806568	& 0.810716 & 0.822861 &	0.822894	& 0.822865 \\ 
& & 512  &  0.924955	& 0.924972	& 0.924956 & 0.804269	& 0.802848 &	0.807301 & 0.823017	 & 0.823042	& 0.823019\\ \cline{2-12} 

& \multirow{3}{*}{Validation} & 128  &  0.908000	 & 0.908000	& 0.908000  & 0.749337 &	0.747838	& 0.751366  & 0.812013 & 	0.812018	& 0.812018\\ 
& & 256 &  0.905160	& 0.905476	& 0.905200  & 0.754998	& 0.753753	& 0.756831 &	0.812903	& 0.812906 &	0.812907\\ 
& & 512  &  0.913603	& 0.913616	& 0.913600 &	0.748105 &	0.747633 &	0.748634 & 0.812691	& 0.812693	& 0.812694 \\ \cline{2-12} 
  
& \multirow{3}{*}{Test} & 128  &  0.909191	& 0.908000	& 0.909200 & 0.785478 &	0.784112 &	0.788251 & 0.815862	& 0.815879	& 0.815862	\\ 
& & 256 & 	0.900796	& 0.901254	& 0.900800  & 0.775626	& 0.774391 &	0.777322 & 0.815974	& 0.815999	& 0.815975 \\ 
& & 512  &  0.911197	& 0.911204	& 0.911200 & 0.777189	& 0.775390	& 0.780055 & 0.815548	& 0.815577	& 0.815549 \\ \hline 

\end{tabular}
\end{table*}

 \begin{table*}[!t]
\caption{Quantitative results ($\mathbf{F1}_w$, $\mathbf{P}_w$, and $\mathbf{R}_w$)  for sentiment analysis using the RoBERTa-BiLSTM model are based on the IMDb, Twitter US Airline, and Sentiment140 datasets} \label{roberta_bilstm_model_p_r_f1}
\centering
\begin{tabular}{p{1cm}|p{1.4cm}|p{1.1cm}||p{.9cm}|p{.9cm}|p{.9cm} ||p{.9cm}|p{.9cm}|p{.9cm}||p{.9cm}|p{.9cm}|p{.9cm}}
\hline \hline

\multirow{2}{*}{\textbf{\makecell{Learning \\Rate ($l$)}}} & \multirow{2}{*}{\textbf{\makecell{Model \\Evaluation}}} & \multirow{2}{*}{\textbf{\makecell{Hidden \\Units ($h$)}}} & \multicolumn{3}{c||}{IMDb Dataset}  & \multicolumn{3}{c||}{Twitter US Airline Dataset} & \multicolumn{3}{c}{Sentiment140 Dataset} \\ \cline{4-12}

 &  &  & \centering $\mathbf{F1}_w$ & \centering $\mathbf{P}_w$ &  \makecell{$\mathbf{R}_w$}  & \centering $\mathbf{F1}_w$ & \centering $\mathbf{P}_w$ & \makecell{$\mathbf{R}_w$}  & \centering $\mathbf{F1}_w$ & \centering $\mathbf{P}_w$ &  \makecell{$\mathbf{R}_w$} \\ \hline \hline

\multirow{9}{*}{0.0001} & \multirow{3}{*}{Training} & 128 &	0.333259	& 0.249933	& 0.499933 & 0.576871 &	0.522566	& 0.675622 &	0.333219 &	0.249897	 & 0.499897 \\ 
& & 256 & 0.835726 & 0.835790  & 0.835733 & 0.652124	& 0.650691	& 0.70636 &	0.333219 &	0.249897	 & 0.499897\\ 
& & 512 &	0.333259	& 0.249933	& 0.499933 &	0.483979	& 0.393857	& 0.627580 &	0.333219 &	0.249897	 & 0.499897 \\ \cline{2-12} 

& \multirow{3}{*}{Validation} & 128 & 0.342260	& 0.258064	& 0.508000 & 0.603579	& 0.562282	& 0.693989 &  0.337725 & 	0.253960 &	0.503944\\ 
& & 256 &	0.834410 & 0.834507 & 0.834400 &	0.659741 &	0.646649 &	0.70765 &  0.337725 & 	0.253960 &	0.503944\\ 
& & 512 &  0.342260	& 0.258064	& 0.508000 & 0.503843 &	0.414018	& 0.643443 &  0.337725 & 	0.253960 &	0.503944\\ \cline{2-12} 
  
& \multirow{3}{*}{Test} & 128 &	 0.325805	& 0.243246 &	0.493200 & 0.552052	 & 0.507510	& 0.654372 & 0.331457	& 0.248312	& 0.498310 \\ 
& & 256 &  0.837603 &  0.837611 & 0.837600 & 0.622737 &	0.609742	& 0.68306 & 0.331457	& 0.248312	& 0.498310\\ 
& & 512 & 0.325805	& 0.243246	& 0.493200 & 0.448003 &	0.358035	& 0.598361 & 0.331457	& 0.248312	& 0.498310 \\ \hline

\multirow{9}{*}{0.00001} & \multirow{3}{*}{Training} & 128  & 0.954020 &	0.954107  &	0.954022 &	0.828071 	& 0.832336 &	0.825668 &	0.861205 &	0.861205 &	0.861205	\\ 
& & 256 &	0.958036 &	0.958392 &	0.958044 & 0.881230	& 0.880769	& 0.882210 &	0.861912 &	0.861914 &	0.861912 \\ 
& & 512 &	93.7617 &	93.7767 &	93.7622 &	0.834221 &	0.834907 &	0.840012 &	0.861940 &	0.861940 &	0.861940 \\ \cline{2-12}

& \multirow{3}{*}{Validation} & 128 &	0.918779 &	0.918942 &	0.918800 &	0.779927 &	0.785797 &	0.775956 &	0.820196 	& 0.820205 	& 0.820194 \\ 
& & 256  &	0.922786 &	0.923801 &	0.922800  & 0.779584	& 0.780695	& 0.780055 &	0.820573 &	0.820604 &	0.820570 \\ 
& & 512 &	0.913549 &	0.914065	& 0.913600 &	0.771039 &	0.767785 &	0.780055 &	0.820208 &	0.820212 &	0.820207 \\ \cline{2-12} 
  
& \multirow{3}{*}{Test} & 128 &	0.918803	& 0.918842	& 0.918800 &	0.784169 &	0.792642 &	0.780055 &	0.822361 &	0.822362 &	0.822361 \\ 
& & 256 & \textbf{0.923529}	& \textbf{0.924562}	& \textbf{0.923600} & \textbf{0.807334}	& \textbf{0.809353}	& \textbf{0.807377} &	\textbf{0.822485} &	\textbf{0.822486} &	\textbf{0.822486}	 \\ 
& & 512 &	0.916802	& 0.917067	& 0.916800 & 0.798464 &	0.797720	& 0.803279 &	0.821984 &	0.821985 &	0.821985  \\ \hline

\multirow{9}{*}{0.000001} & \multirow{3}{*}{Training} & 128  &	0.922776	& 0.922821 &	0.922778 & 0.804605	& 0.803048	& 0.808288 & 0.822227 & 	0.822256	& 0.822230\\ 
& & 256 &  0.922800 &	0.922800	& 0.922800 &	0.806545	& 0.805071	& 0.810033 & 0.822150	& 0.822182 &	0.822154 \\ 
& & 512  &  0.919013	& 0.919213	& 0.919022 &	0.803353 &	0.801858 &	0.806846 & 0.822890	& 0.822913	& 0.822893 \\ \cline{2-12} 

& \multirow{3}{*}{Validation} & 128  & 0.911603 & 	0.911621	& 0.911600 & 0.759718	& 0.757919	& 0.762295 & 0.812627 &	0.812631 &	0.812631  \\ 
& & 256 &  0.910399	& 0.910399	& 0.910400 &	0.750240 & 	0.748403 &	0.752732 &	0.812564 &	0.812569	& 0.812569 \\ 
& & 512  &  0.905163	& 0.905443	& 0.905200 &	0.747260 & 	0.744380	& 0.751366 & 0.812579	& 0.812580	& 0.812581 \\ \cline{2-12} 
  
& \multirow{3}{*}{Test} & 128  &  0.908792 &	0.908826 & 	0.908800 & 0.789076	& 0.787694	& 0.790984 &	0.815298 &	0.815319 &	0.815299	\\ 
& & 256 & 	0.909201	& 0.909205	& 0.909200 &	0.784110	& 0.782377	 & 0.786885 &	0.815573 &	0.815602 &	0.815574 \\ 
& & 512  & 0.906404 &	0.906574	& 0.906400 & 0.781751	& 0.780117 &	0.784153 & 0.815875	& 0.815891	& 0.815874 \\ \hline 

\end{tabular}
\end{table*}

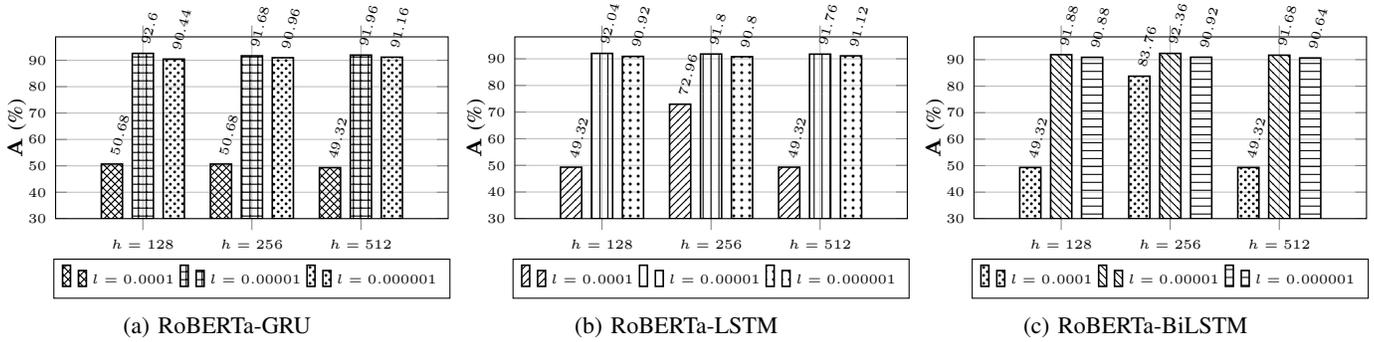
\begin{figure*} [!t]
     \centering
     \begin{subfigure}[b]{0.32\linewidth}
         \captionsetup{justification=centering}
         \begin{tikzpicture}[scale=1]
            \begin{axis}[
    ybar=.13cm,
    every node near coord/.append style={font=\tiny},
    legend style={font=\tiny},
    tick label style={font=\tiny},
    ylabel near ticks, ylabel shift={-6pt},
    width=6.8cm,
    height=4cm,
    every node near coord/.append style={
                        anchor=west,
                        rotate=75
                },
    enlargelimits=.40,
    enlarge y limits={0.1,upper},
    legend style={at={(0.5,-0.22)},
    anchor=north, legend columns=-1},
    ymin=30, 
    ylabel={$\mathbf{A}$ (\%)},
    symbolic x coords={$h=128$, $h=256$, $h=512$},
    xtick=data,
    ytick={10, 20, 30, 40, 50, 60, 70, 80, 90, 100},
    grid=both,
    nodes near coords,
    nodes near coords align={vertical},
    bar width=8pt,
    label style={font=\footnotesize},
    ]
\addplot [draw=black, semithick, pattern=crosshatch, pattern color = black] coordinates {($h=128$,50.68) ($h=256$,50.68) ($h=512$,49.32)};  

\addplot [draw=black, semithick, pattern=grid,  pattern color = black] coordinates {($h=128$,92.60) ($h=256$,91.68) ($h=512$,91.96)};  

\addplot [draw=black, semithick, pattern=crosshatch dots,  pattern color = black] coordinates {($h=128$,90.44) ($h=256$,90.96) ($h=512$,91.16)}; 


\legend{$l=0.0001$, $l=0.00001$, $l=0.000001$}
\end{axis}
 
        \end{tikzpicture}
        \caption{RoBERTa-GRU}
         \label{roberta_gru_accuracy_imdb}
     \end{subfigure}
     \hfill
       \begin{subfigure}[b]{0.32\linewidth}
        \captionsetup{justification=centering}
         \begin{tikzpicture}[scale=1]
            \begin{axis}[
    ybar=.13cm,
    every node near coord/.append style={font=\tiny},
    legend style={font=\tiny},
    tick label style={font=\tiny},
    ylabel near ticks, ylabel shift={-6pt},
    width=6.8cm,
    height=4cm,
    every node near coord/.append style={
                        anchor=west,
                        rotate=75
                },
    enlargelimits=.40,
    enlarge y limits={0.1,upper},
    legend style={at={(0.5,-0.22)},
    anchor=north, legend columns=-1},
    ymin=30, 
    ylabel={$\mathbf{A}$ (\%)},
    symbolic x coords={$h=128$, $h=256$, $h=512$},
    xtick=data,
    ytick={10, 20, 30, 40, 50, 60, 70, 80, 90, 100},
    grid=both,
    nodes near coords,
    nodes near coords align={vertical},
    bar width=8pt,
    label style={font=\footnotesize},
    ]
\addplot [draw=black, semithick, pattern=north east lines, pattern color = black] coordinates {($h=128$,49.32) ($h=256$,72.96) ($h=512$,49.32)};  

\addplot [draw=black, semithick, pattern=vertical lines,  pattern color = black] coordinates {($h=128$,92.04) ($h=256$,91.80) ($h=512$,91.76)};  

\addplot [draw=black, semithick, pattern=dots,  pattern color = black] coordinates {($h=128$,90.92) ($h=256$,90.80) ($h=512$,91.12)}; 


\legend{$l=0.0001$, $l=0.00001$, $l=0.000001$}
\end{axis}
 
        \end{tikzpicture}
        \caption{RoBERTa-LSTM}
         \label{roberta_lstm_accuracy_imdb}
     \end{subfigure}
     \hfill
     \begin{subfigure}[b]{0.32\linewidth}
        \captionsetup{justification=centering}
         \begin{tikzpicture}[scale=1]
            \begin{axis}[
    ybar=.13cm,
    every node near coord/.append style={font=\tiny},
    legend style={font=\tiny},
    tick label style={font=\tiny},
    ylabel near ticks, ylabel shift={-6pt},
    width=6.8cm,
    height=4cm,
    every node near coord/.append style={
                        anchor=west,
                        rotate=75
                },
    enlargelimits=.40,
    enlarge y limits={0.1,upper},
    legend style={at={(0.5,-0.22)},
    anchor=north, legend columns=-1},
    ymin=30, 
    ylabel={$\mathbf{A}$ (\%)},
    symbolic x coords={$h=128$, $h=256$, $h=512$},
    xtick=data,
    ytick={10, 20, 30, 40, 50, 60, 70, 80, 90, 100},
    grid=both,
    nodes near coords,
    nodes near coords align={vertical},
    bar width=8pt,
    label style={font=\footnotesize},
    ]
\addplot [draw=black, semithick, pattern=crosshatch dots, pattern color = black] coordinates {($h=128$,49.32) ($h=256$,83.76) ($h=512$,49.32)};  

\addplot [draw=black, semithick, pattern=north west lines,  pattern color = black] coordinates {($h=128$,91.88) ($h=256$,92.36) ($h=512$,91.68)};  

\addplot [draw=black, semithick, pattern=horizontal lines,  pattern color = black] coordinates {($h=128$,90.88) ($h=256$,90.92) ($h=512$,90.64)}; 


\legend{$l=0.0001$, $l=0.00001$, $l=0.000001$}
\end{axis}
 
        \end{tikzpicture}
       \caption{RoBERTa-BiLSTM}
         \label{roberta_bilstm_accuracy_imdb}
     \end{subfigure}
     \hfill
          \caption{Assessing the test accuracy of the RoBERTa-GRU, RoBERTa-LSTM, and RoBERTa-BiLSTM models using a range of hyperparameters: $l=\{0.0001, 0.00001, 0.000001\}$, $h=\{128, 256, 512\}$. The models are trained for 5 \texttt{epochs} using the \texttt{AdamW} optimizer on the IMDb dataset.}
        \label{roberta_model_accuracy_imdb}
\end{figure*}

\begin{figure*} [!t]
     \centering
     \begin{subfigure}[b]{0.32\linewidth}
         \captionsetup{justification=centering}
         \begin{tikzpicture}[scale=1]
            \begin{axis}[
    ybar=.13cm,
    every node near coord/.append style={font=\tiny},
    legend style={font=\tiny},
    tick label style={font=\tiny},
    ylabel near ticks, ylabel shift={-6pt},
    width=6.8cm,
    height=4cm,
    every node near coord/.append style={
                        anchor=west,
                        rotate=75
                },
    enlargelimits=.40,
    enlarge y limits={0.1,upper},
    legend style={at={(0.5,-0.22)},
    anchor=north, legend columns=-1},
    ymin=30, 
    ylabel={$\mathbf{A}$ (\%)},
    symbolic x coords={$h=128$, $h=256$, $h=512$},
    xtick=data,
    ytick={10, 20, 30, 40, 50, 60, 70, 80, 90, 100},
    grid=both,
    nodes near coords,
    nodes near coords align={vertical},
    bar width=8pt,
    label style={font=\footnotesize},
    ]
\addplot [draw=black, semithick, pattern=crosshatch, pattern color = black] coordinates {($h=128$,63.11) ($h=256$,61.20) ($h=512$,66.94)};  

\addplot [draw=black, semithick, pattern=grid,  pattern color = black] coordinates {($h=128$,79.51) ($h=256$,78.42) ($h=512$,79.92)};  

\addplot [draw=black, semithick, pattern=crosshatch dots,  pattern color = black] coordinates {($h=128$,77.46) ($h=256$,77.46) ($h=512$,77.73)}; 


\legend{$l=0.0001$, $l=0.00001$, $l=0.000001$}
\end{axis}
 
        \end{tikzpicture}
        \caption{RoBERTa-GRU}
         \label{roberta_gru_accuracy_twitter}
     \end{subfigure}
     \hfill
       \begin{subfigure}[b]{0.32\linewidth}
        \captionsetup{justification=centering}
         \begin{tikzpicture}[scale=1]
            \begin{axis}[
    ybar=.13cm,
    every node near coord/.append style={font=\tiny},
    legend style={font=\tiny},
    tick label style={font=\tiny},
    ylabel near ticks, ylabel shift={-6pt},
    width=6.8cm,
    height=4cm,
    every node near coord/.append style={
                        anchor=west,
                        rotate=75
                },
    enlargelimits=.40,
    enlarge y limits={0.1,upper},
    legend style={at={(0.5,-0.22)},
    anchor=north, legend columns=-1},
    ymin=30, 
    ylabel={$\mathbf{A}$ (\%)},
    symbolic x coords={$h=128$, $h=256$, $h=512$},
    xtick=data,
    ytick={10, 20, 30, 40, 50, 60, 70, 80, 90, 100},
    grid=both,
    nodes near coords,
    nodes near coords align={vertical},
    bar width=8pt,
    label style={font=\footnotesize},
    ]
\addplot [draw=black, semithick, pattern=north east lines, pattern color = black] coordinates {($h=128$,59.84) ($h=256$,69.13) ($h=512$,67.76)};  

\addplot [draw=black, semithick, pattern=vertical lines,  pattern color = black] coordinates {($h=128$,79.10) ($h=256$,80.33) ($h=512$,80.05)};  

\addplot [draw=black, semithick, pattern=dots,  pattern color = black] coordinates {($h=128$,78.83) ($h=256$,77.73) ($h=512$,78.01)}; 


\legend{$l=0.0001$, $l=0.00001$, $l=0.000001$}
\end{axis}
 
        \end{tikzpicture}
        \caption{RoBERTa-LSTM}
         \label{roberta_lstm_accuracy_twitter}
     \end{subfigure}
     \hfill
     \begin{subfigure}[b]{0.32\linewidth}
        \captionsetup{justification=centering}
         \begin{tikzpicture}[scale=1]
            \begin{axis}[
    ybar=.13cm,
    every node near coord/.append style={font=\tiny},
    legend style={font=\tiny},
    tick label style={font=\tiny},
    ylabel near ticks, ylabel shift={-6pt},
    width=6.8cm,
    height=4cm,
    every node near coord/.append style={
                        anchor=west,
                        rotate=75
                },
    enlargelimits=.40,
    enlarge y limits={0.1,upper},
    legend style={at={(0.5,-0.22)},
    anchor=north, legend columns=-1},
    ymin=30, 
    ylabel={$\mathbf{A}$ (\%)},
    symbolic x coords={$h=128$, $h=256$, $h=512$},
    xtick=data,
    ytick={10, 20, 30, 40, 50, 60, 70, 80, 90, 100},
    grid=both,
    nodes near coords,
    nodes near coords align={vertical},
    bar width=8pt,
    label style={font=\footnotesize},
    ]
\addplot [draw=black, semithick, pattern=crosshatch dots, pattern color = black] coordinates {($h=128$,65.44) ($h=256$,68.31) ($h=512$,59.84)};  

\addplot [draw=black, semithick, pattern=north west lines,  pattern color = black] coordinates {($h=128$,78.01) ($h=256$,80.74) ($h=512$,80.33)};  

\addplot [draw=black, semithick, pattern=horizontal lines,  pattern color = black] coordinates {($h=128$,79.10) ($h=256$,78.69) ($h=512$,78.42)}; 


\legend{$l=0.0001$, $l=0.00001$, $l=0.000001$}
\end{axis}
 
        \end{tikzpicture}
       \caption{RoBERTa-BiLSTM}
         \label{roberta_bilstm_accuracy_twitter}
     \end{subfigure}
     \hfill
          \caption{Assessing the test accuracy of the RoBERTa-GRU, RoBERTa-LSTM, and RoBERTa-BiLSTM models using a range of hyperparameters: $l=\{0.0001, 0.00001, 0.000001\}$, $h=\{128, 256, 512\}$. The models are trained for 5 \texttt{epochs} using the \texttt{AdamW} optimizer on the Twitter US Airline dataset.}
        \label{roberta_model_accuracy_twitter}
\end{figure*}
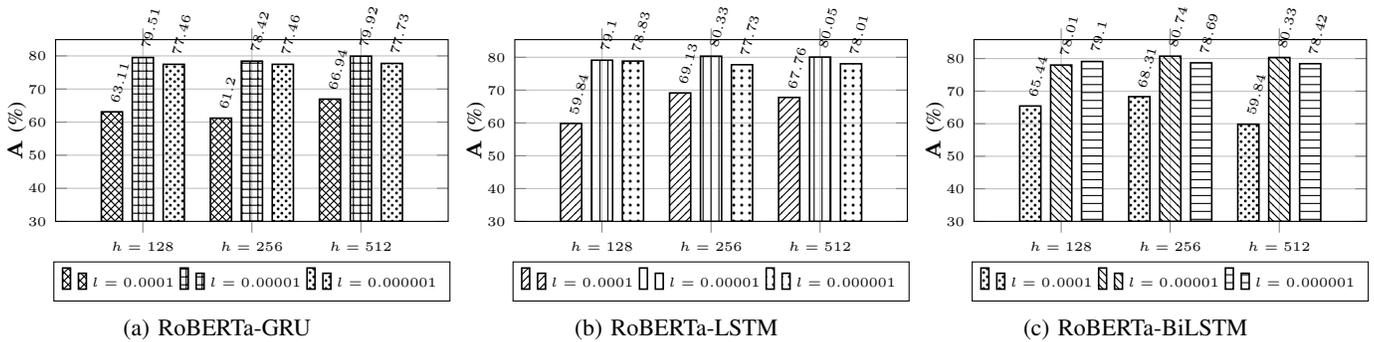

\begin{figure*} [!t]
     \centering
     \begin{subfigure}[b]{0.32\linewidth}
         \captionsetup{justification=centering}
         \begin{tikzpicture}[scale=1]
            \begin{axis}[
    ybar=.13cm,
    every node near coord/.append style={font=\tiny},
    legend style={font=\tiny},
    tick label style={font=\tiny},
    ylabel near ticks, ylabel shift={-6pt},
    width=6.8cm,
    height=4cm,
    every node near coord/.append style={
                        anchor=west,
                        rotate=75
                },
    enlargelimits=.40,
    enlarge y limits={0.1,upper},
    legend style={at={(0.5,-0.22)},
    anchor=north, legend columns=-1},
    ymin=30, 
    ylabel={$\mathbf{A}$ (\%)},
    symbolic x coords={$h=128$, $h=256$, $h=512$},
    xtick=data,
    ytick={10, 20, 30, 40, 50, 60, 70, 80, 90, 100},
    grid=both,
    nodes near coords,
    nodes near coords align={vertical},
    bar width=8pt,
    label style={font=\footnotesize},
    ]
\addplot [draw=black, semithick, pattern=crosshatch, pattern color = black] coordinates {($h=128$,49.83) ($h=256$,49.83) ($h=512$,49.83)};  

\addplot [draw=black, semithick, pattern=grid,  pattern color = black] coordinates {($h=128$,82.25) ($h=256$,82.32) ($h=512$,82.21)};  

\addplot [draw=black, semithick, pattern=crosshatch dots,  pattern color = black] coordinates {($h=128$,81.62) ($h=256$,81.62) ($h=512$,81.63)}; 


\legend{$l=0.0001$, $l=0.00001$, $l=0.000001$}
\end{axis}
 
        \end{tikzpicture}
        \caption{RoBERTa-GRU}
         \label{roberta_gru_accuracy_sentiment140}
     \end{subfigure}
     \hfill
       \begin{subfigure}[b]{0.32\linewidth}
        \captionsetup{justification=centering}
         \begin{tikzpicture}[scale=1]
            \begin{axis}[
    ybar=.13cm,
    every node near coord/.append style={font=\tiny},
    legend style={font=\tiny},
    tick label style={font=\tiny},
    ylabel near ticks, ylabel shift={-6pt},
    width=6.8cm,
    height=4cm,
    every node near coord/.append style={
                        anchor=west,
                        rotate=75
                },
    enlargelimits=.40,
    enlarge y limits={0.1,upper},
    legend style={at={(0.5,-0.22)},
    anchor=north, legend columns=-1},
    ymin=30, 
    ylabel={$\mathbf{A}$ (\%)},
    symbolic x coords={$h=128$, $h=256$, $h=512$},
    xtick=data,
    ytick={10, 20, 30, 40, 50, 60, 70, 80, 90, 100},
    grid=both,
    nodes near coords,
    nodes near coords align={vertical},
    bar width=8pt,
    label style={font=\footnotesize},
    ]
\addplot [draw=black, semithick, pattern=north east lines, pattern color = black] coordinates {($h=128$,49.83) ($h=256$,49.83) ($h=512$,49.83)};  

\addplot [draw=black, semithick, pattern=vertical lines,  pattern color = black] coordinates {($h=128$,82.26) ($h=256$,82.14) ($h=512$,82.29)};  

\addplot [draw=black, semithick, pattern=dots,  pattern color = black] coordinates {($h=128$,81.59) ($h=256$,81.60) ($h=512$,81.55)}; 


\legend{$l=0.0001$, $l=0.00001$, $l=0.000001$}
\end{axis}
 
        \end{tikzpicture}
        \caption{RoBERTa-LSTM}
         \label{roberta_lstm_accuracy_sentiment140}
     \end{subfigure}
     \hfill
     \begin{subfigure}[b]{0.32\linewidth}
        \captionsetup{justification=centering}
         \begin{tikzpicture}[scale=1]
            \begin{axis}[
    ybar=.13cm,
    every node near coord/.append style={font=\tiny},
    legend style={font=\tiny},
    tick label style={font=\tiny},
    ylabel near ticks, ylabel shift={-6pt},
    width=6.8cm,
    height=4cm,
    every node near coord/.append style={
                        anchor=west,
                        rotate=75
                },
    enlargelimits=.40,
    enlarge y limits={0.1,upper},
    legend style={at={(0.5,-0.22)},
    anchor=north, legend columns=-1},
    ymin=30, 
    ylabel={$\mathbf{A}$ (\%)},
    symbolic x coords={$h=128$, $h=256$, $h=512$},
    xtick=data,
    ytick={10, 20, 30, 40, 50, 60, 70, 80, 90, 100},
    grid=both,
    nodes near coords,
    nodes near coords align={vertical},
    bar width=8pt,
    label style={font=\footnotesize},
    ]
\addplot [draw=black, semithick, pattern=crosshatch dots, pattern color = black] coordinates {($h=128$,49.83) ($h=256$,49.83) ($h=512$,49.83)};  

\addplot [draw=black, semithick, pattern=north west lines,  pattern color = black] coordinates {($h=128$,82.24) ($h=256$,82.25) ($h=512$,82.20)};  

\addplot [draw=black, semithick, pattern=horizontal lines,  pattern color = black] coordinates {($h=128$,81.53) ($h=256$,81.56) ($h=512$,81.59)}; 


\legend{$l=0.0001$, $l=0.00001$, $l=0.000001$}
\end{axis}
 
        \end{tikzpicture}
       \caption{RoBERTa-BiLSTM}
         \label{roberta_bilstm_accuracy_sentiment140}
     \end{subfigure}
     \hfill
          \caption{Assessing the test accuracy of the RoBERTa-GRU, RoBERTa-LSTM, and RoBERTa-BiLSTM models using a range of hyperparameters: $l=\{0.0001, 0.00001, 0.000001\}$, $h=\{128, 256, 512\}$. The models are trained for 5 \texttt{epochs} using the \texttt{AdamW} optimizer on the Sentiment140 dataset.}
        \label{roberta_model_accuracy_sentiment140}
\end{figure*}
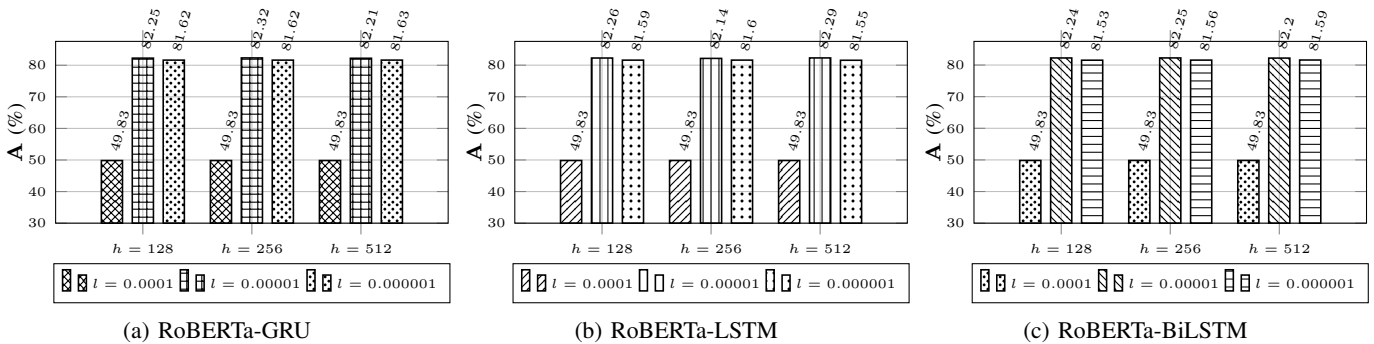

Tables \ref{roberta_gru_model_p_r_f1}-\ref{roberta_bilstm_model_p_r_f1} display the results of $\mathbf{F1}_w$, $\mathbf{P}_w$, and $\mathbf{R}_w$ metrics for the RoBERTa-GRU, RoBERTa-LSTM, and RoBERTa-BiLSTM models across three datasets. These experiments adhere to specific hyperparameter settings (i.e., $l=\{0.0001, 0.00001,0.000001\}$, $epoch=5$, $h=\{128, 256, 512\}$, $d=0.1$, and optimizer=\texttt{AdamW}). Evaluation encompasses all models across training, validation, and test data splits. The RoBERTa-GRU model achieved $\mathbf{F1}_w$, $\mathbf{P}_w$, and $\mathbf{R}_w$  scores of 92.60\%, 92.64\%, and 92.60\%, respectively, for the IMDb; and 79.06\%, 79.42\%, and 79.92\%, respectively, for the Twitter US Airline. It also garnered an $\mathbf{F1}_w$, $\mathbf{P}_w$, and $\mathbf{R}_w$  score of 82.32\% for the Sentiment140 dataset. The RoBERTa-LSTM model obtained $\mathbf{F1}_w$, $\mathbf{P}_w$, and $\mathbf{R}_w$ scores of 92.04\% for the IMDb; 80.32\%, 80.47\%, and 80.33\%, respectively, for the Twitter US Airline; and  $\mathbf{F1}_w$, $\mathbf{P}_w$, and $\mathbf{R}_w$ scores of 82.29\% for the Sentiment140 dataset. On the other hand, the RoBERTa-BiLSTM model achieved $\mathbf{F1}_w$, $\mathbf{P}_w$, and $\mathbf{R}_w$ scores of 92.35\%, 92.46\%, and 92.36\%, respectively, for the IMDb; 80.73\%, 80.94\%, and 80.74\%, respectively, for the Twitter US Airline; and 82.25\% for the Sentiment140 dataset with $l=0.00001$ and $h=256$. Furthermore, RoBERTa-BiLSTM obtained superior $\mathbf{F1}_w$, $\mathbf{P}_w$, and $\mathbf{R}_w$ scores of 95.80\%, 95.84\%, and 95.80\%, respectively, during training, and 92.28\%, 92.38\%, and 92.28\%, respectively, during validation on the Sentiment140 dataset when compared to other models. It can be seen that the RoBERTa-GRU and RoBERTa-LSTM models failed to outperform the RoBERTa-BiLSTM model on the IMDb and Twitter US Airline datasets. However, the RoBERTa-GRU model exhibited marginally better performance (with $\mathbf{F1}_w$, $\mathbf{P}_w$, and $\mathbf{R}_w$ scores of 82.32\% for the Test data split) on the Sentiment140 dataset compared to the others models.

\begin{table*}[h]
\caption{Quantitative results for sentiment analysis employing the RoBERTa-GRU, RoBERTa-LSTM, and RoBERTa-BiLSTM models with three distinct optimizers (\texttt{SGD}, \texttt{RMSprop}, and \texttt{Rprop}), alongside fixed hyperparameters of $l=0.00001$ and $h=256$, are presented. The models are trained for 5 \texttt{epochs} on the IMDb, Twitter US Airline, and Sentiment140 datasets.} \label{roberta_models_optimizers_p_r_f11}
\centering
\begin{tabular}{p{1.1cm}|p{2.1cm}|p{1.1cm}||p{.9cm}|p{.9cm}|p{.9cm} ||p{.9cm}|p{.9cm}|p{.9cm}||p{.9cm}|p{.9cm}|p{.9cm}}
\hline \hline

\multirow{2}{*}{\textbf{\makecell{Optimizer}}} & \multirow{2}{*}{\textbf{\makecell{Model}}} & \multirow{2}{*}{\textbf{\makecell{Model \\ Evaluation}}} & \multicolumn{3}{c||}{IMDb Dataset}  & \multicolumn{3}{c||}{Twitter US Airline Dataset} & \multicolumn{3}{c}{Sentiment140 Dataset} \\ \cline{4-12}

 &  &  & \centering $\mathbf{F1}_w$ & \centering $\mathbf{P}_w$ &  \makecell{$\mathbf{R}_w$}  & \centering $\mathbf{F1}_w$ & \centering $\mathbf{P}_w$ & \makecell{$\mathbf{R}_w$}  & \centering $\mathbf{F1}_w$ & \centering $\mathbf{P}_w$ &  \makecell{$\mathbf{R}_w$} \\ \hline \hline

\multirow{9}{*}{SGD} & \multirow{3}{*}{RoBERTa-LSTM} & Training &	0.963533 &	0.963570 &	0.963533 & 0.484001 &	0.393887 &	0.627580 &	0.746607 &	0.746608 &	0.746607 \\ 
& & Validation & 0.916404 &	0.916431 & 	0.916400 & 0.503843 &	0.414018 &	0.643443 &	0.747410 &	0.747415 &	0.747408\\ 
& & Test & 0.920796 &	0.920810 &	0.920800 & 0.448003 &	0.358035 &	0.598361 &	0.748623 &	0.748630 &	0.748623 \\ \cline{2-12} 

& \multirow{3}{*}{RoBERTa-BiLSTM} & Training &	0.965733 &	0.965738 &	0.965733 & 0.483979 &	0.393857 	& 0.627580 &	0.741888 &	0.741962 &	0.741904 \\ 
& & Validation & 0.925197 &	0.925202 &	0.925200 & 0.503843 &	0.414018 &	0.643443 &	0.741821 &	0.741940 &	0.741824\\ 
& & Test & 0.929595 &	0.929620 &	0.929600 & 0.448003 & 	0.358035 &	0.598361 &	0.742817 & 	0.742878 &	0.742838 \\ \cline{2-12}

& \multirow{3}{*}{RoBERTa-GRU} & Training &		0.626660 &	0.637177 &	0.630911 & 0.483979 &	0.393857 &	0.627580 &	0.745768 &	0.746113 &	0.745840 \\ 
& & Validation & 0.628920 &	0.637516 &	0.633200 & 0.503843 & 	0.414018 &	0.643443 &	0.745893 &	0.746358 &	0.745943\\ 
& & Test &  0.612301 &	0.626618 &	0.617600  & 0.448003 &	0.358035 &	0.598361 &	0.745891 &	0.746187 &	0.745968 \\ \hline

\multirow{9}{*}{RMSprop} & \multirow{3}{*}{RoBERTa-LSTM} & Training &	0.953308 &	0.954296 &	0.953333 & 0.882130 &	0.883383 &	0.881603 &	0.835982 &	0.836355 &	0.836021 \\ 
& & Validation & 0.915163 &	0.916821 &	0.915200 & 0.792620 &	0.800576 &	0.789617 &	0.812310 & 	0.812576 & 0.812369\\ 
& & Test & 0.919888 &	0.921577 &	0.920000 & 0.797753 &	0.806394 &	0.795082 &	0.814887 &	0.815251 &	0.814923 \\ \cline{2-12} 

& \multirow{3}{*}{RoBERTa-BiLSTM} & Training &	0.964042 &	0.964182 &	0.964044 & 0.881943 &	0.881552 &	0.882438 &	0.837697 &	0.837741 & 	0.837702 \\ 
& & Validation & 0.924805 &	0.924905 &	0.924800 & 0.784245 & 	0.785119 &	0.784153 &	0.813109 &	0.813181 &	0.813107\\ 
& & Test & 0.927978 &	0.928223 &	0.928000 & 0.798651 &	0.801249 &	0.797814 &	0.817244 &	0.817276 &	0.817252 \\ \cline{2-12}

& \multirow{3}{*}{RoBERTa-GRU} & Training &		0.958806 &	0.959552 &	0.958822 & 0.885970 &	0.886035 &	0.885929 &	0.823524 &	0.823648 &	0.823538
 \\ 
& & Validation & 0.919167 &	0.920749 &	0.919200 & 0.783879 &	0.783694 &	0.784153 &	0.805222 &	0.805287 &	0.805244\\ 
& & Test & 0.920689 &	0.922380 &	0.920800 & 0.809253 &	0.814673 &	0.806011 &	0.808515 &	0.808637 &	0.808524 \\ \hline

\multirow{9}{*}{Rprop} & \multirow{3}{*}{RoBERTa-LSTM} & Training &		0.962152 & 	0.962310 &	0.962156 & 0.815864 &	0.814367 &	0.820431 &	0.795564 &	0.795577 &	0.795566 \\ 
& & Validation & 0.917604 &	0.917839 &	0.917600 & 0.775373 &	0.772986 &	0.781421 &	0.790234 &	0.790249 &	0.790231\\ 
& & Test & 0.919178 &	0.919386 & 	0.919200 & 0.781915 &	0.780306 &	0.785519 &	0.792521 & 	0.792522 &	0.792522 \\ \cline{2-12} 

& \multirow{3}{*}{RoBERTa-BiLSTM} & Training &	0.928581 &	0.929063 &	0.9286 & 0.907253 &	0.907184 &	0.908925 &	0.822915 &	0.823134 &	0.822941 \\ 
& & Validation & 0.917997 &	0.918589 
 & 0.9180 & 0.781882 &	0.779601 &	0.786885 &	0.804620 &	0.804749 &	0.804655\\ 
& & Test &  0.909949 &	0.910488 	& 0.9100 & 0.798375 &	0.798201 &	0.800546 &	0.806506 &	0.806782 &	0.806534 \\ \cline{2-12}

& \multirow{3}{*}{RoBERTa-GRU} & Training &		0.968666 &	0.968710 &	0.968667 & 0.926327 &	0.926314 &	0.926457 &	0.795433 &	0.795617 &	0.795461 \\ 
& & Validation & 0.917605 &	0.917779 &	0.917600 & 0.801109 &	0.801719 &	0.800546 &	0.789173 &	0.789467 &	0.789192\\ 
& & Test & 0.919581 &	0.919746 &	0.919600 & 0.801475 &	0.804779 &	0.799180 &	0.791183 &	0.791363 &	0.791220 \\ \hline

\end{tabular}
\end{table*}

Figures \ref{roberta_model_accuracy_imdb}-\ref{roberta_model_accuracy_sentiment140} illustrate the accuracy of the RoBERTa-GRU, RoBERTa-LSTM, and RoBERTa-BiLSTM models with various hyperparameters across three datasets. The RoBERTa-GRU model achieved a higher $\mathbf{A}$ score of 92.60\% with $l=0.00001$ and $h=128$ for the IMDb dataset, 79.92\% with $l=0.00001$ and $h=512$ for the Twitter US Airline dataset, and 82.32\% with $l=0.00001$ and $h=256$ for the Sentiment140 dataset. Similarly, the RoBERTa-LSTM model obtained a higher $\mathbf{A}$ scores of 92.04\%, 80.33\%, and 82.29\% for the IMDb, Twitter US Airline, and Sentiment140 datasets, respectively, with $l=0.00001$ and $h$ values of 128, 256, and 512. In contrast, the RoBERTa-BiLSTM model achieved a superior $\mathbf{A}$ scores of 92.36\%, 80.74\%, and 82.25\% for the IMDb, Twitter US Airline, and Sentiment140 datasets, respectively, with $l=0.00001$ and $h=256$. It is evident that the RoBERTa-GRU and RoBERTa-LSTM models attained higher $\mathbf{A}$ scores with different $h$ values (128, 256, and 512) across datasets. On the other hand, the RoBERTa-BiLSTM model consistently yielded better $\mathbf{A}$ scores across datasets when utilizing $l=0.00001$ and $h=256$. The RoBERTa-BiLSTM model outperformed other models when employing the hyperparameter settings $l=0.00001$ and $h=256$ across datasets.

Moreover, the impact of different optimizers on model performance is explored, as detailed in Table \ref{roberta_models_optimizers_p_r_f11}. Three additional optimizers—\texttt{SGD}, \texttt{RMSprop}, and \texttt{Rprop} \cite{choi2019empirical}—are examined, each with a learning rate ($l=0.00001$) and hidden units ($h=256$), showcasing their effects across datasets. The RoBERTa-BiLSTM model achieved the highest $\mathbf{F1}_w$ scores of 92.96\% and 92.80\% using \texttt{SGD} and \texttt{RMSprop} optimizers, respectively, for the IMDb dataset. Meanwhile, the RoBERTa-GRU model obtained a slightly higher $\mathbf{F1}_w$ score of 91.96\% compared to the RoBERTa-BiLSTM model (which achieved an $\mathbf{F1}_w$ score of 91.00\%) when employing the \texttt{Rprop} optimizer. For the Twitter US Airline dataset, all models yielded suboptimal results with \texttt{SGD}. Among them, the RoBERTa-BiLSTM model obtained higher $\mathbf{F1}_w$, $\mathbf{P}_w$, and $\mathbf{R}_w$ scores of 44.80\%, 35.80\%, and 59.84\%, respectively.
However, notable enhancements are observed with \texttt{RMSprop} and \texttt{Rprop} optimizers, where the RoBERTa-GRU model achieves the $\mathbf{F1}_w$ scores of 80.93\% and 80.15\%, respectively. On the other hand, across the Sentiment140 dataset, all models exhibit nearly identical results with \texttt{SGD}, hovering around $\mathbf{F1}_w$, $\mathbf{P}_w$, and $\mathbf{R}_w$ scores of approximately 74.50\%. Yet, the RoBERTa-BiLSTM model demonstrates better performance, achieving higher $\mathbf{F1}_w$ scores of 81.72\% and 80.65\% with \texttt{RMSprop} and \texttt{Rprop}, respectively, compared to other models. It is clear that in most cases, the models yield inferior results when utilizing these alternative optimizers (\texttt{SGD}, \texttt{RMSprop}, and \texttt{Rprop}), in comparison to those attained with  \texttt{AdamW} optimizer.

 \begin{table*}[h]
\caption{Quantitative results for sentiment analysis using the BERT-GRU, BERT-LSTM, and BERT-BiLSTM models with the hyperparameter set $l= 0.00001$, $h= 256$, and optimizer= \texttt{AdamW}. The models are trained for 5 \texttt{epochs} on the IMDb, Twitter US Airline, and Sentiment140 datasets.} \label{bert_model_p_r_f1}
\centering
\begin{tabular}{p{1.7cm}|p{1.4cm}||p{.9cm}|p{.9cm}|p{.9cm} ||p{.9cm}|p{.9cm}|p{.9cm}||p{.9cm}|p{.9cm}|p{.9cm}}
\hline \hline

\multirow{2}{*}{\textbf{\makecell{Model}}} & \multirow{2}{*}{\textbf{\makecell{Model \\Evaluation}}}  & \multicolumn{3}{c||}{IMDb Dataset}  & \multicolumn{3}{c||}{Twitter US Airline Dataset} & \multicolumn{3}{c}{Sentiment140 Dataset} \\ \cline{3-11}

&  & \centering $\mathbf{F1}_w$ & \centering $\mathbf{P}_w$ &  \makecell{$\mathbf{R}_w$}  & \centering $\mathbf{F1}_w$ & \centering $\mathbf{P}_w$ & \makecell{$\mathbf{R}_w$}  & \centering $\mathbf{F1}_w$ & \centering $\mathbf{P}_w$ &  \makecell{$\mathbf{R}_w$} \\ \hline \hline

\multirow{3}{*}{BERT-GRU} & Training & 0.959173	& 0.959407 &	0.959178 & 0.809176	& 0.807872 &	0.812234 & 0.857501	& 0.857591 &	0.857509 \\ \cline{2-11}
& Validation & 	0.906006 & 	0.906140 &	0.906000 &	0.756455 &	0.754155	& 0.759563 & 0.814906 &	0.815074	& 0.814910 \\ \cline{2-11}
& Test & 	0.911175 &	0.911380 &	0.911200 & 	0.775682	& 0.774644 &	0.777322 & 0.818300	& 0.818387	& 0.818316 \\ \hline

\multirow{3}{*}{BERT-LSTM} & Training & 0.957062 &	0.957256 &	0.957067 & 0.804269	& 0.802848 &	0.807301 &	0.857269	& 0.857271	& 0.857269  \\ \cline{2-11}
& Validation & 0.908406 &	0.908540 &	0.908400 &	0.748105 &	0.747633 &	0.748634 &	0.815563 &	0.815568	& 0.815561   \\ \cline{2-11}
& Test & 	0.913163	& 0.913528	& 0.913200 & 0.777189	& 0.775390	& 0.780055 &	0.817501	 & 0.817503	& 0.817502 \\ \hline

\multirow{3}{*}{BERT-BiLSTM} & Training &  0.958131 &	0.958214	& 0.958133 &	0.803353 &	0.801858 &	0.806846 & 0.856565	& 0.856565	& 0.856565 \\ \cline{2-11}
& Validation & 	0.910402	& 0.910409 &	0.910400 &	0.747260 & 	0.744380	& 0.751366 & 0.814562 & 	0.814568	& 0.814560  \\ \cline{2-11}
& Test & 	0.912382	& 0.912518 &	0.912400  & 0.781751	& 0.780117 &	0.784153 &	0.818091	& 0.818090 &	0.818091 \\ \hline

\end{tabular}
\end{table*}

\begin{figure} [h]
     \centering
    
         \captionsetup{justification=centering}
         \begin{tikzpicture}[scale=1]
            \begin{axis}[
    ybar=.13cm,
    every node near coord/.append style={font=\tiny},
    legend style={font=\tiny},
    tick label style={font=\tiny},
    ylabel near ticks, ylabel shift={-6pt},
    width=\linewidth,
    height=5cm,
    every node near coord/.append style={
                        anchor=west,
                        rotate=75
                },
    enlargelimits=.40,
    enlarge y limits={0.1,upper},
    legend style={at={(0.5,-0.22)},
    anchor=north, legend columns=-1},
    ymin=30, 
    ylabel={$\mathbf{A}$ (\%)},
    symbolic x coords={IMDb, Twitter US Airline, Sentiment140},
    xtick=data,
    ytick={10, 20, 30, 40, 50, 60, 70, 80, 90, 100},
    grid=both,
    nodes near coords,
    nodes near coords align={vertical},
    bar width=12pt,
    label style={font=\footnotesize},
    ]
\addplot [draw=teal, semithick, pattern=horizontal lines, pattern color = teal] coordinates {(IMDb,91.12) (Twitter US Airline,78.83) (Sentiment140,81.83)};  

\addplot [draw=blue, semithick, pattern=bricks,  pattern color = blue] coordinates {(IMDb,91.32) (Twitter US Airline,78.83) (Sentiment140,81.75)};  

\addplot [draw=magenta, semithick, pattern=crosshatch dots,  pattern color = magenta] coordinates {(IMDb,91.24) (Twitter US Airline,79.37) (Sentiment140,81.81)}; 


\legend{BERT-GRU, BERT-LSTM, BERT-BiLSTM}
\end{axis}
 
        \end{tikzpicture}     
    
          \caption{Assessing the test accuracy of the BERT-GRU, BERT-LSTM, BERT-BiLSTM models using the hyperparameter set $l= 0.00001$ and $h=256$. The models are trained for 5 \texttt{epochs} using the \texttt{AdamW} optimizer.}
        \label{berta_base_accuracy_imdb_twitter_sentment140}
\end{figure}
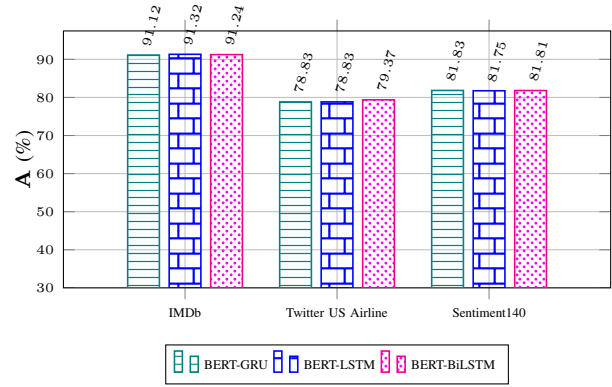

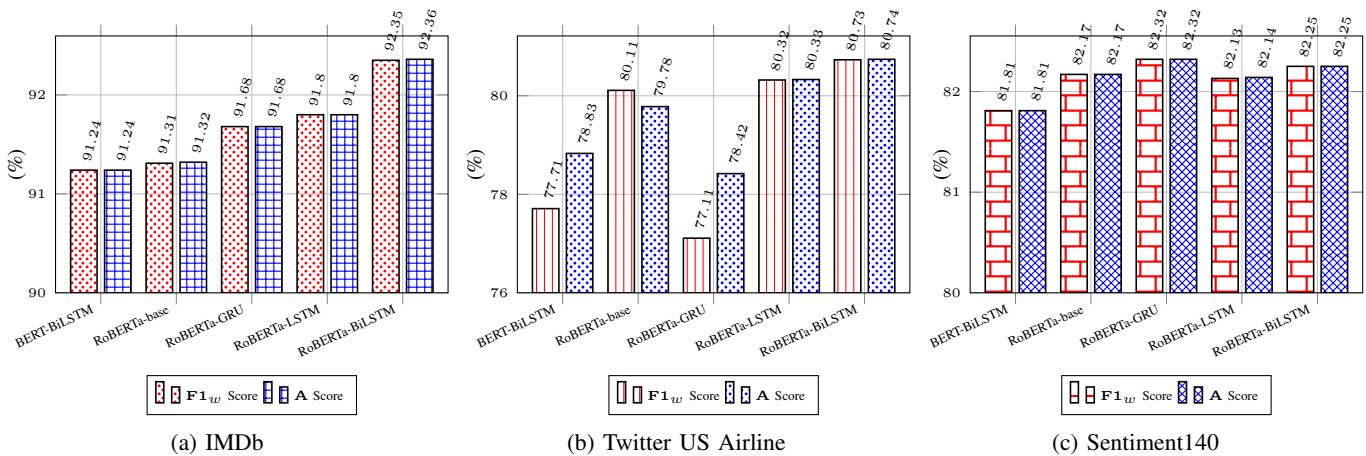
\begin{figure*} [!t]
     \centering
     \begin{subfigure}[b]{0.32\linewidth}
         \captionsetup{justification=centering}
         \begin{tikzpicture}[scale=1]
            \begin{axis}[
    ybar=.1cm,
    every node near coord/.append style={font=\tiny},
    legend style={font=\tiny},
    tick label style={font=\tiny},
    ylabel near ticks, ylabel shift={-6pt},
    width=6.8cm,
    height=5cm,
    every node near coord/.append style={
                        anchor=west,
                        rotate=75
                },
    enlargelimits=.15,
    enlarge y limits={0.1,upper},
    legend style={at={(0.5,-0.32)},
    anchor=north, legend columns=-1},
    ymin=90, 
    ylabel={(\%)},
    symbolic x coords={ BERT-BiLSTM, RoBERTa-base, RoBERTa-GRU, RoBERTa-LSTM, RoBERTa-BiLSTM},
    xtick=data,
    x tick label style={rotate=25,anchor=east},
    grid=both,
    nodes near coords,
    nodes near coords align={vertical},
    bar width=10pt,
    label style={font=\footnotesize},
    ]
\addplot [draw=black, semithick, pattern=crosshatch dots, pattern color = red] coordinates { (BERT-BiLSTM,91.24) (RoBERTa-base,91.31) (RoBERTa-GRU,91.68) (RoBERTa-LSTM,91.80) (RoBERTa-BiLSTM,92.35)};  

\addplot [draw=black, semithick, pattern=grid,  pattern color = blue] coordinates {(BERT-BiLSTM,91.24) (RoBERTa-base,91.32) (RoBERTa-GRU,91.68) (RoBERTa-LSTM,91.80) (RoBERTa-BiLSTM,92.36)};  


\legend{$\mathbf{F1}_w$ Score, $\mathbf{A}$ Score}
\end{axis}
 
        \end{tikzpicture}
        \caption{IMDb}
         \label{performance_comparison_imdb}
     \end{subfigure}
     \hfill
       \begin{subfigure}[b]{0.32\linewidth}
        \captionsetup{justification=centering}
         \begin{tikzpicture}[scale=1]
            \begin{axis}[
    ybar=.1cm,
    every node near coord/.append style={font=\tiny},
    legend style={font=\tiny},
    tick label style={font=\tiny},
    ylabel near ticks, ylabel shift={-5pt},
    width=6.8cm,
    height=5cm,
    every node near coord/.append style={
                        anchor=west,
                        rotate=75
                },
    enlargelimits=.15,
    enlarge y limits={0.1,upper},
    legend style={at={(0.5,-0.32)},
    anchor=north, legend columns=-1},
    ymin=76, 
    ylabel={(\%)},
    symbolic x coords={ BERT-BiLSTM, RoBERTa-base, RoBERTa-GRU, RoBERTa-LSTM, RoBERTa-BiLSTM},
    xtick=data,
    x tick label style={rotate=25,anchor=east},
    grid=both,
    nodes near coords,
    nodes near coords align={vertical},
    bar width=10pt,
    label style={font=\footnotesize},
    ]
\addplot [draw=black, semithick, pattern=vertical lines, pattern color = red] coordinates { (BERT-BiLSTM,77.71) (RoBERTa-base,80.11) (RoBERTa-GRU,77.11) (RoBERTa-LSTM,80.32) (RoBERTa-BiLSTM,80.73)};  

\addplot [draw=black, semithick, pattern=crosshatch dots,  pattern color = blue] coordinates { (BERT-BiLSTM,78.83) (RoBERTa-base,79.78) (RoBERTa-GRU,78.42) (RoBERTa-LSTM,80.33) (RoBERTa-BiLSTM,80.74)};  


\legend{$\mathbf{F1}_w$ Score, $\mathbf{A}$ Score}
\end{axis}
 
        \end{tikzpicture}
        \caption{Twitter US Airline}
         \label{performance_comparison_twitter}
     \end{subfigure}
     \hfill
     \begin{subfigure}[b]{0.32\linewidth}
        \captionsetup{justification=centering}
         \begin{tikzpicture}[scale=1]
            \begin{axis}[
    ybar=.1cm,
    every node near coord/.append style={font=\tiny},
    legend style={font=\tiny},
    tick label style={font=\tiny},
    ylabel near ticks, ylabel shift={-10pt},
    width=6.8cm,
    height=5cm,
    every node near coord/.append style={
                        anchor=west,
                        rotate=75
                },
    enlargelimits=.15,
    enlarge y limits={0.1,upper},
    legend style={at={(0.5,-0.32)},
    anchor=north, legend columns=-1},
    ymin=80, 
    ylabel={(\%)},
    symbolic x coords={ BERT-BiLSTM, RoBERTa-base, RoBERTa-GRU, RoBERTa-LSTM, RoBERTa-BiLSTM},
    xtick=data,
    ytick={80, 81,82,83},
    x tick label style={rotate=25,anchor=east},
    grid=both,
    nodes near coords,
    nodes near coords align={vertical},
    bar width=10pt,
    label style={font=\footnotesize},
    ]
\addplot [draw=black, semithick, pattern=bricks, pattern color = red] coordinates { (BERT-BiLSTM,81.81) (RoBERTa-base,82.17) (RoBERTa-GRU,82.32) (RoBERTa-LSTM,82.13) (RoBERTa-BiLSTM,82.25)};  

\addplot [draw=black, semithick, pattern=crosshatch,  pattern color = blue] coordinates { (BERT-BiLSTM,81.81) (RoBERTa-base,82.17) (RoBERTa-GRU,82.32) (RoBERTa-LSTM,82.14) (RoBERTa-BiLSTM,82.25)};  


\legend{$\mathbf{F1}_w$ Score, $\mathbf{A}$ Score}
\end{axis}
 
        \end{tikzpicture}
       \caption{Sentiment140}
         \label{performance_comparison_sentiment140}
     \end{subfigure}
     \hfill
          \caption{Comparisons of  $\mathbf{F1}_w$ and  $\mathbf{A}$ scores among BERT-BiLSTM, RoBERTa-base, RoBERTa-GRU, RoBERTa-LSTM, and RoBERTa-BiLSTM models, considering hyperparameters $l = 0.00001$, $h = 256$, and optimizer=\texttt{AdamW} across the IMDb, Twitter US Airline, and Sentiment140 datasets.}
        \label{performance_comparisons_bert_roberta}
\end{figure*}

Furthermore, experiments are conducted using the BERT-GRU, BERT-LSTM, and BERT-BiLSTM models with identical hyperparameter settings (i.e., $l=0.00001$, $h=256$, optimizer=\texttt{AdamW}) across datasets. Table \ref{bert_model_p_r_f1} illustrates that the BERT-GRU model achieved $\mathbf{F1}_w$ scores of 91.11\%, 77.57\%, and 81.83\% for the IMDb, Twitter US Airline, and Sentiment140 datasets, respectively, when evaluated with the test data splits. Similarly, the BERT-LSTM model garnered $\mathbf{F1}_w$ scores of 91.32\%, 77.72\%, and 81.75\% for the same datasets. In contrast, the BERT-BiLSTM model obtained $\mathbf{F1}_w$ scores of 91.24\%, 78.18\%, and 81.81\% for the respective datasets. Figure \ref{berta_base_accuracy_imdb_twitter_sentment140} presents a comparative analysis of test accuracy among the BERT-based models. The BERT-BiLSTM model achieved higher $\mathbf{A}$ of 91.24\%, 79.37\%, and 81.81\% for the IMDb, Twitter US Airline, and Sentiment140 datasets, respectively. The BERT-BiLSTM model performed relatively better compared to others. It consistently garnered higher results across all datasets. However, none of the BERT models achieved superior results compared to any of the RoBERTa-based models.

\begin{table*}[h]
    \centering
    \caption{The experimental results ($\mathbf{P}$, $\mathbf{R}$, $\mathbf{F1}$, and $\mathbf{A}$ ) comparisons between ML models and the proposed RoBERTa-BiLSTM model for sentiment analysis on the IMDb, Twitter US Airline, and Sentiment140 datasets.}
    \label{comparison_ML_roberta}
    \begin{tabular}{c||c|c|c|c||c|c|c|c||c|c|c|c}
    \hline
         \multirow{2}{*}{\textbf{ ML Models}} & \multicolumn{4}{c||}{\textbf{IMDb Dataset}}  & \multicolumn{4}{c||}{\textbf{Twitter US Airline Dataset}}  & \multicolumn{4}{c}{\textbf{Sentimet140 Dataset}}   \\ \cline{2-13}
          & \textbf{P} & \textbf{R} & \textbf{F1} & \textbf{A} & \textbf{P} & \textbf{R} & \textbf{F1} & \textbf{A} & \textbf{P} & \textbf{R} & \textbf{F1} & \textbf{A} \\ \hline \hline
         
         NB \cite{jung2016enhanced} & {0.87} & {0.87} & {0.87} & 0.8701 &  {0.79} & {0.44} & {0.45} & 0.6950 & {0.77} & {0.77} & {0.77} & 0.7657 \\ \hline
         
         LR \cite{dholpuria2018sentiment} & 0.90 & 0.90 & 0.90 & 0.8712 &  0.78 & 0.69 & 0.72 & 0.8050 & 0.78 & 0.78 & 0.78 & 0.7801 \\ \hline
         
         DT \cite{zharmagambetov2015sentiment} &  0.74 & 0.73 & 0.73 & 0.7346 & 0.62 & 0.56 & 0.58  & 0.7114 & 0.69 & 0.62 & 0.59 & 0.6234 \\ \hline
         
        KNN \cite{dholpuria2018sentiment} &  0.78 & 0.77 & 0.77 & 0.7737 & 0.60 & 0.60 & 0.60 & 0.6841 & 0.66 & 0.60 & 0.57 & 0.6039\\ \hline
         
         AdaBoost \cite{vadivukarassi2018exploration} &  0.83 & 0.83 & 0.83 & 0.8337 & 0.67 & 0.63 & 0.65 & 0.7459 & 0.71 & 0.70 & 0.69 & 0.6994\\ \hline \hline
         
         \textbf{RoBERTa-BiLSTM} &  \textbf{0.9246} & \textbf{0.9236} & \textbf{0.9235} & \textbf{0.9236} & \textbf{0.8094 } & \textbf{0.8074} & \textbf{0.8073} & \textbf{0.8074} & \textbf{ 0.8225} & \textbf{0.8225 } & \textbf{0.8225} & \textbf{0.8225}\\ 
         
         \hline
    \end{tabular}
\end{table*}

A comparison among the best-performing models including BERT-BiLSTM, RoBERTa-base, RoBERTa-GRU, RoBERTa-LSTM, and RoBERTa-BiLSTM is conducted, considering hyperparameters $l = 0.00001$, $h = 256$, and optimizer=\texttt{AdamW} across the IMDb, Twitter US Airline, and Sentiment140 datasets, as illustrated in Figure \ref{performance_comparisons_bert_roberta}. Figure \ref{performance_comparison_imdb} demonstrates that the proposed RoBERTa-BiLSTM model achieves the highest $\mathbf{F1}_w$ and $\mathbf{A}$ scores of 92.35\% and 92.36\%, respectively, surpassing other top-performing models on the IMDb dataset. Similarly, for the Twitter US Airline dataset, the RoBERTa-BiLSTM model attains $\mathbf{F1}_w$ and $\mathbf{A}$ scores of 80.73\% and 80.74\%, respectively (Figure \ref{performance_comparison_twitter}), enhancing classification performance by approximately 0.40\% compared to the nearest best model, RoBERTa-LSTM ($\mathbf{F1}_w$ and $\mathbf{A}$ scores of 80.32\% and 80.33\%, respectively). On the other hand, for the Sentiment140 dataset, the RoBERTa-GRU model achieves $\mathbf{F1}_w$ and $\mathbf{A}$ scores of 82.32\% (Figure \ref{performance_comparison_sentiment140} ), while the RoBERTa-BiLSTM model achieves $\mathbf{F1}_w$ and $\mathbf{A}$ scores of 82.25\%, slightly lower (by 0.07\%) than the best model performance.

\begin{table*}[h]
    \centering
    \caption{The experimental results ($\mathbf{P}$, $\mathbf{R}$, $\mathbf{F1}$, and $\mathbf{A}$ ) comparisons between DL models and the proposed RoBERTa-BiLSTM model for sentiment analysis on the IMDb, Twitter US Airline, and Sentiment140 datasets.}
    \label{comparison_DL_roberta}
    \begin{tabular}{c||c|c|c|c||c|c|c|c||c|c|c|c}
    \hline
         \multirow{2}{*}{\textbf{ DL Models}} & \multicolumn{4}{c||}{\textbf{IMDb Dataset}}  & \multicolumn{4}{c||}{\textbf{Twitter US Airline Dataset}}  & \multicolumn{4}{c}{\textbf{Sentimet140 Dataset}}   \\ \cline{2-13}
          & \textbf{P} & \textbf{R} & \textbf{F1} & \textbf{A} & \textbf{P} & \textbf{R} & \textbf{F1} & \textbf{A} & \textbf{P} & \textbf{R} & \textbf{F1} & \textbf{A} \\ \hline \hline
         
         GRU \cite{hossen2021hotel} & {0.88} & {0.88} & {0.88} & {0.8788} &  {0.73} & {0.71} & {0.72} & {0.7855} & {0.78} & {0.78} & {0.78} & {0.7896} \\ \hline
         
         LSTM \cite{hossen2021hotel} & 0.85 & 0.85 & 0.85 & {0.8511} &  0.71 & 0.69 & 0.69 & {0.7756} & 0.79 & 0.79 & 0.79 & {0.7910}\\ \hline
         
         BiLSTM \cite{garg2020psent20} &  0.87 & 0.86 & 0.86 & {0.8628} & 0.71 & 0.69 & 0.70 & {0.7746} & 0.78 & 0.78 & 0.78 & {0.7853}\\ \hline
         
        CNN-LSTM \cite{jain2021hybrid} &  0.86 & 0.86 & 0.86 & {0.88607} & 0.68 & 0.69 & 0.69 & {0.7602} & 0.77 & 0.77 & 0.77 & {0.7753}\\ \hline

        CNN-BiLSTM \cite{rhanoui2019cnn} &  0.86 & 0.86 & 0.86 & {0.8616} & 0.70 & 0.65 & 0.67 & {0.7732} & 0.77 & 0.77 & 0.77 & {0.7758}\\ \hline \hline

         \textbf{RoBERTa-BiLSTM} &  \textbf{0.9246} & \textbf{0.9236} & \textbf{0.9235} & \textbf{0.9236} & \textbf{0.8094 } & \textbf{0.8074} & \textbf{0.8073} & \textbf{0.8074} & \textbf{ 0.8225} & \textbf{0.8225 } & \textbf{0.8225} & \textbf{0.8225}\\ 
         
         \hline
    \end{tabular}
\end{table*}

\section{Discussion} \label{discussion}

In this section, we discuss the performance of the proposed RoBERTa-BiLSTM model and conduct a comparative analysis with various ML and DL methods within the realm of sentiment analysis. The discussion addresses the impact of data augmentation on model performance for imbalanced datasets. Additionally, we examine the scalability and limitations of the proposed model.

\subsection{Performance Analysis}

This paper introduces a hybrid approach combining LLM and RNN for sentiment analysis. We conduct comprehensive experiments using various models, including BERT, RoBERTa, RoBERTa-GRU, RoBERTa-LSTM, and RoBERTa-BiLSTM, with different hyperparameter sets (such as learning rate ($l$), optimizers, hidden RNN units ($h$)), across three datasets: IMDb, Twitter US Airline, and Sentiment140. Tables \ref{roberta_base_model_p_r_f1}-\ref{roberta_models_optimizers_p_r_f11} present the $\mathbf{F1}_w$, $\mathbf{P}_w$, and $\mathbf{R}_w$ scores obtained by these models across datasets. Among these models, RoBERTa-BiLSTM achieved the highest $\mathbf{F1}_w$ scores of 92.35\%, 80.73\%, and 82.25\% for the IMDb, Twitter, and Sentiment140 datasets, respectively, surpassing the performance of RoBERTa-base, RoBERTa-GRU, and RoBERTa-LSTM. Furthermore, the experiments encompass various BERT-based models, including BERT-GRU, BERT-LSTM, and BERT-BiLSTM, as presented in Table \ref{bert_model_p_r_f1}. It is evident that none of the BERT models outperforms the RoBERTa-based models on any dataset. Figure \ref{performance_comparisons_bert_roberta} presents a comparative analysis of the experimented models, showcasing the effectiveness of the RoBERTa-BiLSTM model across all three datasets by consistently achieving top results.

To ensure a fair comparison with the proposed RoBERTa-BiLSTM model, we consider the performance of both ML and DL models for sentiment analysis on the IMDb, Twitter, and Sentiment140 datasets. Table \ref{comparison_ML_roberta} displays the performance of ML models such as NB, LR, DT, AdaBoost, KNN, alongside RoBERTa-BiLSTM on these datasets. Among the ML models, LR \cite{dholpuria2018sentiment} achieved the highest $\mathbf{A}$ and $\mathbf{F1}$ scores of 87.12\% and 90.00\%, respectively, on the IMDb dataset. However, the proposed RoBERTa-BiLSTM model surpassed these results with an $\mathbf{A}$ and $\mathbf{F1}$ score of 92.36\% and 92.35\%, respectively, marking an improvement of approximately 5.00\% in $\mathbf{A}$ and 2.35\% in $\mathbf{F1}$ score over the LR model. Similarly, on the Twitter dataset, LR \cite{dholpuria2018sentiment} achieved the top $\mathbf{A}$ and $\mathbf{F1}$ score of 80.50\% and 72.00\%, respectively, among ML models. Nevertheless, these scores are lower (about 0.25\% in $\mathbf{A}$ and 8.00\% in $\mathbf{F1}$ score) compared to the RoBERTa-BiLSTM model. For the Sentiment140 dataset, again LR \cite{dholpuria2018sentiment} attained the highest $\mathbf{A}$ and $\mathbf{F1}$ scores of 78.01\% and 78.00\%, respectively, among other ML models. 
In contrast, the RoBERTa-BiLSTM model attained $\mathbf{A}$ and $\mathbf{F1}$ scores of 82.25\%, marking a notable improvement of approximately 4.00\% in both $\mathbf{A}$ and $\mathbf{F1}$ scores compared to the ML models.

Table \ref{comparison_DL_roberta} illustrates the performance of DL models for the sentiment analysis task across the same datasets. The GRU \cite{hossen2021hotel} model achieved higher $\mathbf{A}$ and $\mathbf{F1}$ scores of 87.88\% and 88.00\%, respectively, for the IMDb dataset, and 78.55\% and 72.00\%, respectively, for the Twitter dataset. In comparison, the proposed RoBERTa-BiLSTM models attained $\mathbf{A}$ and $\mathbf{F1}$ scores of 92.36\% and 92.35\%, respectively, for the IMDb dataset, and 80.74\% and 80.73\%, respectively, for the Twitter dataset. The proposed model demonstrated improvements of approximately 5.00\% in $\mathbf{A}$ and $\mathbf{F1}$ scores for the IMDb dataset and about 2.20\% in $\mathbf{A}$ and 8.00\% in $\mathbf{F1}$ scores for the Twitter US Airline dataset compared to the GRU \cite{hossen2021hotel} model. Conversely, the LSTM \cite{hossen2021hotel} model achieved $\mathbf{A}$ and $\mathbf{F1}$ scores of 79.10\% and 79.00\% for the Sentiment140 dataset, while the RoBERTa-BiLSTM model obtained $\mathbf{A}$ and $\mathbf{F1}$ scores of 82.25\%, marking an enhancement of 3.25\% in $\mathbf{A}$ and $\mathbf{F1}$ scores compared to LSTM \cite{hossen2021hotel}.

\begin{figure} [h]
\captionsetup{justification=centering}
     \centering
     \begin{subfigure}[b]{0.23\textwidth}
 \centering
  \includegraphics[width=.9\linewidth]{Tweets.png}
  \caption{Before Augmentation}
   \label{before_aug_twitter_us_airline}
     \end{subfigure}
     \hfill
       \begin{subfigure}[b]{0.23\textwidth}
     \centering
  \includegraphics[width=1\linewidth]{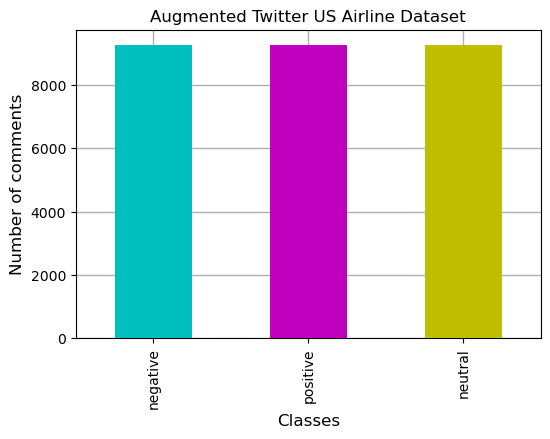}
  \caption{After Augmentation}
   \label{after_aug_twitter_us_airline}
     \end{subfigure}
     \hfill
       \caption{Twitter US Airline dataset before and after data augmentation.}
        \label{augm_main_datasets}
\end{figure}

The pretrained RoBERTa model, having been trained on vast amounts of text data, possesses the ability to discern intricate patterns and relationships between words and phrases. Additionally, its dynamic masking patterns enable it to generalize and adapt to new text sequences. RoBERTa encodes lengthy text sequences into word embedding representations, while the BiLSTM model excels at capturing long-distance dependencies within the input by processing it in both forward and backward directions. The proposed RoBERTa-BiLSTM model capitalizes the power of both RoBERTa and BiLSTM models. Experimental results and comparisons highlight the efficacy of the RoBERTa-BiLSTM model in the sentiment analysis task.

\subsection{Impact of Data Augmentation}

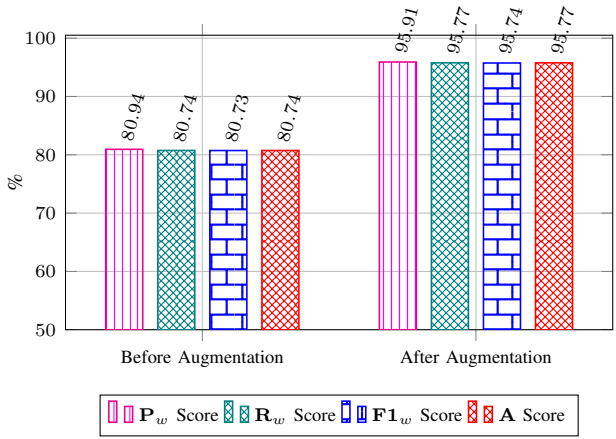
\begin{figure} 
     \centering
    
         \captionsetup{justification=centering}
         \begin{tikzpicture}[scale=1]
            \begin{axis}[
    ybar=.2cm,
    every node near coord/.append style={font=\scriptsize},
    legend style={font=\scriptsize},
    tick label style={font=\scriptsize},
    ylabel near ticks, ylabel shift={-6pt},
    width=\linewidth,
    height=5.5cm,
    every node near coord/.append style={
                        anchor=west,
                        rotate=75
                },
    enlargelimits=.50,
    enlarge y limits={0.1,upper},
    legend style={at={(0.5,-0.22)},
    anchor=north, legend columns=-1},
    ymin=50, 
    ylabel={\%},
    symbolic x coords={Before Augmentation, After Augmentation},
    xtick=data,
    nodes near coords,
    ytick={10, 20, 30, 40, 50, 60, 70, 80, 90, 100},
    grid=both,
    nodes near coords align={vertical},
    bar width=14pt,
    label style={font=\footnotesize},
    ]

\addplot [draw=magenta, semithick, pattern=vertical lines, pattern color = magenta] coordinates {(Before Augmentation,80.94) (After Augmentation,95.91) };  

\addplot [draw=teal, semithick, pattern=crosshatch, pattern color = teal] coordinates {(Before Augmentation,80.74) (After Augmentation,95.77) };  

\addplot [draw=blue, semithick, pattern=bricks, pattern color = blue] coordinates {(Before Augmentation,80.73) (After Augmentation,95.74) };  

\addplot [draw=red, semithick, pattern=crosshatch,  pattern color = red] coordinates {(Before Augmentation,80.74) (After Augmentation,95.77)};  



\legend{$\mathbf{P}_w$ Score, $\mathbf{R}_w$ Score, $\mathbf{F1}_w$ Score, $\mathbf{A}$ Score}
\end{axis}
 
        \end{tikzpicture}     
    
          \caption{Comparing the performance of the RoBERTa-BiLSTM model before and after data augmentation on the Twitter US Airline dataset}
        \label{augmentation_roberta_bilstm}
\end{figure}

Data augmentation aims to balance the class samples by increasing the sample size of minority classes \cite{augmentation9515019}. In this study, the Twitter US Airline dataset exhibits class imbalance, as depicted in Figure \ref{twitter_us_airline}. The objective of data augmentation is to evaluate the performance of the RoBERTa-BiLSTM model before and after augmentation. Figure \ref{augm_main_datasets} depicts the Twitter US Airline dataset before and after augmentation. We synthetically generate samples for the neutral and positive classes to match the sample count of the negative class. Figure \ref{augmentation_roberta_bilstm} compares the performance of the RoBERTa-BiLSTM model on the Twitter US Airline dataset before and after augmentation. Post-augmentation, the model achieves $\mathbf{F1}_w$ and $\mathbf{A}$  scores of 95.74\% and 95.77\%, respectively, representing a notable improvement of approximately 15\% in both $\mathbf{A}$  and $\mathbf{F1}_w$  scores. Thus, augmenting the imbalanced dataset significantly enhances the model's performance.

\subsection{Scalability}

Sentiment analysis is a text analytical application used to discern the polarity of text/comments by accurately interpreting the underlying meaning of the provided text. The proposed RoBERTa-BiLSTM model showcases its efficacy in sentiment analysis tasks by achieving higher $\mathbf{A}$ scores (92.36\%, 80.74\%, and 82.25\% for the IMDb, Twitter, and Sentiment140 datasets, respectively) in comment classification. Additionally, the proposed model notably enhances the average $\mathbf{A}$ score of 0.70\% compared to RoBERTa-base model. These findings suggest that the proposed RoBERTa-BiLSTM model holds potential for various application domains such as business, economics, politics, education, and programming. 

The concept behind the proposed hybrid architecture can be adapted for various task-specific LLMs such as CodeBERT, CodeT5, CodeT5+, and Llama. Particularly, in programming education, the fine-tuned model can be effectively employed for tasks like programming language identification, error detection in program code, code-clone detection, code generation, and code summarization. Additionally, the architecture of RoBERTa-BiLSTM can be beneficial in other application domains where tasks involve complex, diverse, and large datasets.

\subsection{Threats to Validity}

In this paper, we proposed a RoBERTa-BiLSTM model for sentiment analysis, encompassing various procedures from data preprocessing to model development. Our proposed model achieved significant results in sentiment analysis compared to other state-of-the-art models across the IMDb, Twitter US Airline, and Sentiment140 datasets. However, the results of the proposed model may vary due to several factors: ($i$) differing strategies for data preprocessing, ($ii$) variations in experimental environments/platforms, ($iii$) diverse hyperparameters and their values, ($iv$) discrepancies in datasets, ($v$) variances in base models (e.g., \texttt{roberta-base}, \texttt{xlm-roberta-base}) of the LLMs, and ($vi$) differences in model architectures.

In the follow-up work, we aim to validate the performance of the proposed RoBERTa-BiLSTM model by addressing the aforementioned challenges.

\section{Conclusion} \label{conclusion_research}

In this paper, we introduce a novel hybrid model, RoBERTa-BiLSTM, designed for the task of sentiment analysis. We provide a detailed description of the architecture and theory underlying the proposed RoBERTa-BiLSTM model. This model leverages the strengths of both RoBERTa and BiLSTM models to effectively analyze text. The proposed model achieves high $\mathbf{F1}$ and $\mathbf{A}$ scores in sentiment analysis tasks across the IMDb, Twitter US Airline, and Sentiment140 datasets. Experimental results demonstrate that the proposed RoBERTa-BiLSTM model attains an average $\mathbf{A}$ and $\mathbf{F1}_w$ score of 85.12\% and 85.11\%, respectively, across all three datasets. In comparison, the RoBERTa-base, RoBERTa-GRU, and RoBERTa-LSTM models achieve average $\mathbf{A}$ of 84.42\%, 84.14\%, and 84.76\%, respectively, along with $\mathbf{F1}_w$ scores of 84.53\%, 83.70\%, and 84.75\%, respectively. Notably, the proposed RoBERTa-BiLSTM model improves sentiment analysis an average $\mathbf{A}$ by 0.70\% compared to the RoBERTa-base model and by 0.36\% compared to the RoBERTa-LSTM model. The proposed model demonstrates superior performance on imbalanced datasets, such as Twitter US Airline. Moreover, we fine-tuned various hyperparameters to assess their impact on model performance. Additionally, we explored the suitability and scalability of the proposed model across other application domains. The integration of RoBERTa and BiLSTM proves to be a powerful, steady, and efficient approach for sentiment analysis, positioning the proposed model as a potential candidate for various NLP tasks.


\bibliographystyle{IEEEtran}



\begin{IEEEbiography}[{\includegraphics[width=1in,height=1.25in,clip,keepaspectratio]{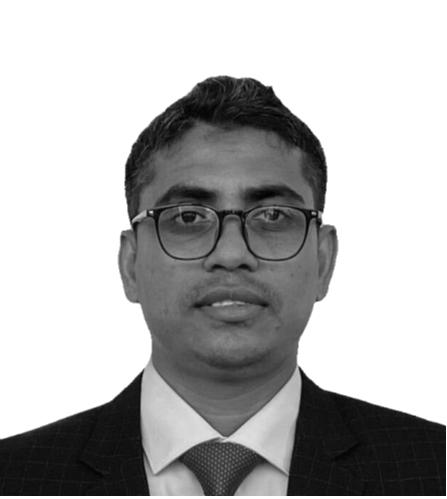}}]{Md. Mostafizer Rahman} received his Ph.D. degree in the Department of Computer and Information Systems, University of Aizu, Japan in 2022. He is also working at Dhaka University of Engineering \& Technology, Gazipur, Bangladesh. He received his B.Sc. and M.Sc. engineering degrees in the Department of Computer Science and Engineering from Hajee Mohammad Danesh Science \& Technology University, Dinajpur, Bangladesh, and Dhaka University of Engineering \& Technology, Gazipur, Bangladesh, in 2009 and 2014, respectively. His research interests include machine learning, LLM, software engineering, AI for Code, NLP, information visualization, and big data analytics.
\end{IEEEbiography}

\begin{IEEEbiography}[{\includegraphics[width=1.in,height=2in,clip,keepaspectratio]{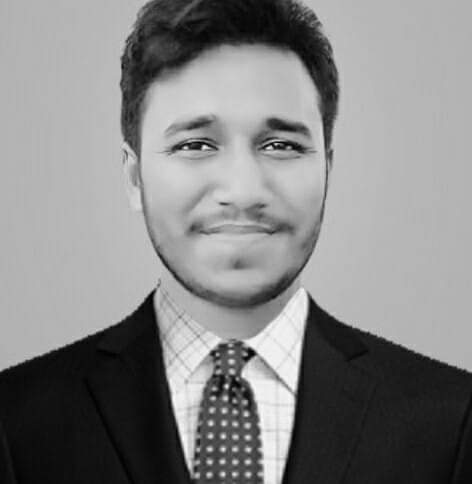}}]{Ariful Islam Shiplu} is a undergraduate student at Dhaka University of Engineering \& Technology, Gazipur, Bangladesh. He completed his Diploma-in-Engineering degree in the Department of Computer Science and Technology from Narsingdi Polytechnic Institute, Narsingdi, Bangladesh. His academic journey is complemented by a robust foundation in technical concepts and adept problem-solving skills. His research interests include Machine Learning, Deep Learning, Large Language Models, and Programming.
\end{IEEEbiography}

\begin{IEEEbiography}[{\includegraphics[width=1in,height=1.25in,clip,keepaspectratio]{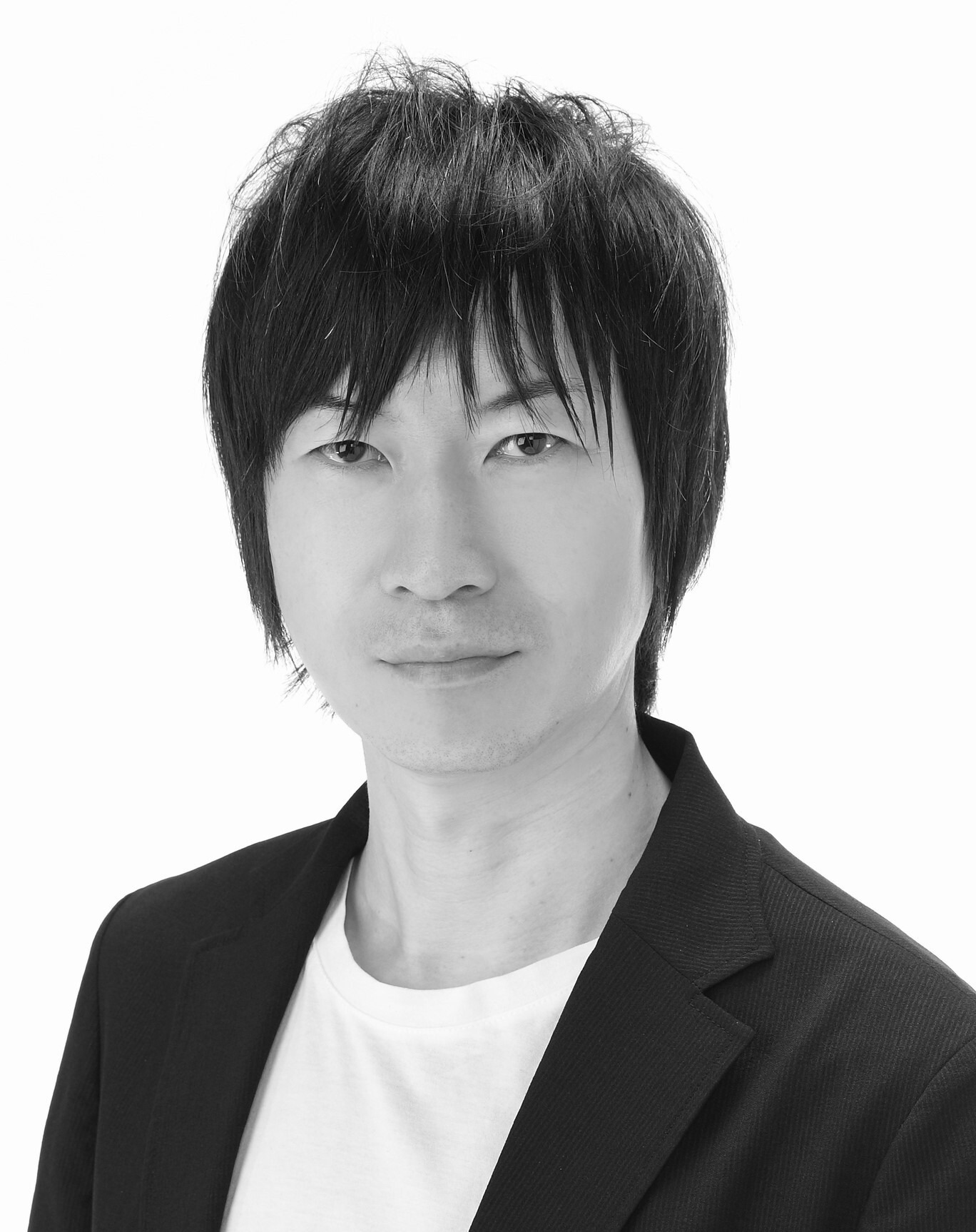}}]{Yutaka Watanobe} is currently a Senior Associate Professor at the School of Computer Science and Engineering, The University of Aizu, Japan. He received his M.S. and Ph.D. degrees from The University of Aizu in 2004 and 2007 respectively. He was a Research Fellow of the Japan Society for the Promotion of Science (JSPS) at The University of Aizu in 2007. He is now a director of i-SOMET. He was a coach of four ICPC World Final teams. He is a developer of the Aizu Online Judge (AOJ) system. His research interests include intelligent software, programming environment, smart learning, machine learning, data mining, cloud robotics, and visual languages. He is a member of IEEE, IPSJ.
\end{IEEEbiography}

\begin{IEEEbiography}[{\includegraphics[width=1in,height=1.25in,clip,keepaspectratio]{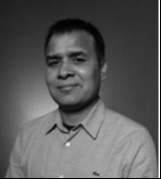}}]{Md Ashad Alam, PhD} is Statistical Scientist (Data Scientist/Biostatistician/Bioinformatician) with over a decade of experience conducting multi- and inter-disciplinary research. His expertise lies in human imaging, genetics/genomics, functional genomics (transcriptomics, proteomics, epigenomics), biostatistics \& bioinformatics, big data \& statistical machine learning, genetic epidemiology, and biomedical data science. The focus of his work is on the integrated analysis of multi-view data in biomedical applications. Dr. Alam has extensive experience in the theoretical development of novel statistical and machine/deep learning methods for gene/protein expression analyses and integrative analyses of various omics data at DNA, mRNA, miRNA, methylation, and protein levels. His work is represented by ~45 peer-reviewed publications (including > 25 in the past 5 years), several of them in top journals. He is highly knowledgeable, experienced, and competent in bioinformatic and statistical issues related to next-generation DNA sequencing-based studies. 
\end{IEEEbiography}

\vfill

\end{document}